\documentclass[smallcondensed,natbib]{svjour3}
\usepackage{mathptmx}
\usepackage{amsmath}
\usepackage{amssymb}
\usepackage{natbib}
\usepackage{graphicx}
\usepackage[raggedright,hang]{subfigure}
\usepackage{algorithmic}
\usepackage{algorithm}
\usepackage{placeins}

\journalname{Machine Learning}

\begin{document}


\title{Finding Anomalous Periodic Time Series}
\subtitle{An Application to Catalogs of Periodic Variable Stars}

\author{Umaa Rebbapragada \and Pavlos Protopapas \and Carla E. Brodley \and Charles Alcock}

\institute{ Umaa Rebbapragada and Carla E. Brodley \at 
       Department of Computer Science\\
       161 College Ave.\\
       Tufts University\\
       Medford, MA  02155 \\
       \email{ \{urebbapr,brodley\}@cs.tufts.edu }\\
       \and 
       Pavlos Protopapas and Charles Alcock \at 
       Harvard-Smithsonian Center for Astrophysics\\
       60 Garden Street\\
       Cambridge, MA 02138 \\
       \email{ \{pprotopapas,calcock\}@cfa.harvard.edu } \\
       \and
       Pavlos Protopapas \at
       Initiative in Innovative Computing\\
       Harvard University\\
       60 Oxford Street\\
       Cambridge, MA 02138
       }

\date{Received: date / Revised version: date}       

\maketitle

\begin{abstract}
Catalogs of periodic variable stars contain large numbers of periodic
light-curves (photometric time series data from the astrophysics
domain).  Separating anomalous objects from well-known classes is an
important step towards the discovery of new classes of astronomical
objects.  Most anomaly detection methods for time series data assume
either a single continuous time series or a set of time series whose
periods are aligned.  Light-curve data precludes the use of these
methods as the periods of any given pair of light-curves may be out of
sync.  One may use an existing anomaly detection method if, prior to
similarity calculation, one performs the costly act of aligning two
light-curves, an operation that scales poorly to massive data sets.
This paper presents PCAD, an unsupervised anomaly detection method for
large sets of unsynchronized periodic time-series data, that outputs a
ranked list of both global and local anomalies. It calculates its
anomaly score for each light-curve in relation to a set of centroids
produced by a modified k-means clustering algorithm.  Our method is
able to scale to large data sets through the use of sampling. We
validate our method on both light-curve data and other time series
data sets.  We demonstrate its effectiveness at finding known
anomalies, and discuss the effect of sample size and number of
centroids on our results.  We compare our method to naive solutions
and existing time series anomaly detection methods for unphased data,
and show that PCAD's reported anomalies are comparable to or better
than all other methods. Finally, astrophysicists on our team have
verified that PCAD finds true anomalies that might be indicative of
novel astrophysical phenomena.  \keywords{Anomaly detection \and Time
  Series Data}
\end{abstract}

\section{Introduction}
\label{sec:Introduction}

Quasars \citep{schmidt1963}, radio pulsars \citep{hewish1968}, and
cosmic gamma-ray bursts \citep{klebesadel1973} were all discovered by
alert scientists who, while examining data for a primary purpose,
encountered aberrant phenomena whose further study led to these
legendary discoveries. Such discoveries were possible in an era when
scientists had a close connection with their data. The advent
of massive data sets renders unexpected discoveries through manual
inspection improbable if not impossible. Fortunately, automated
anomaly detection programs may resurrect this mode of discovery and
identify atypical phenomena indicative of novel astronomical objects.

\begin{figure*}[ptb]

    \centering
    \mbox{
    \subfigure[Cepheid. OGLE-LMC\_SC4-53463. Period 5.4 days.] {
           \label{fig:typical-ceph}
       \includegraphics[width=2in]{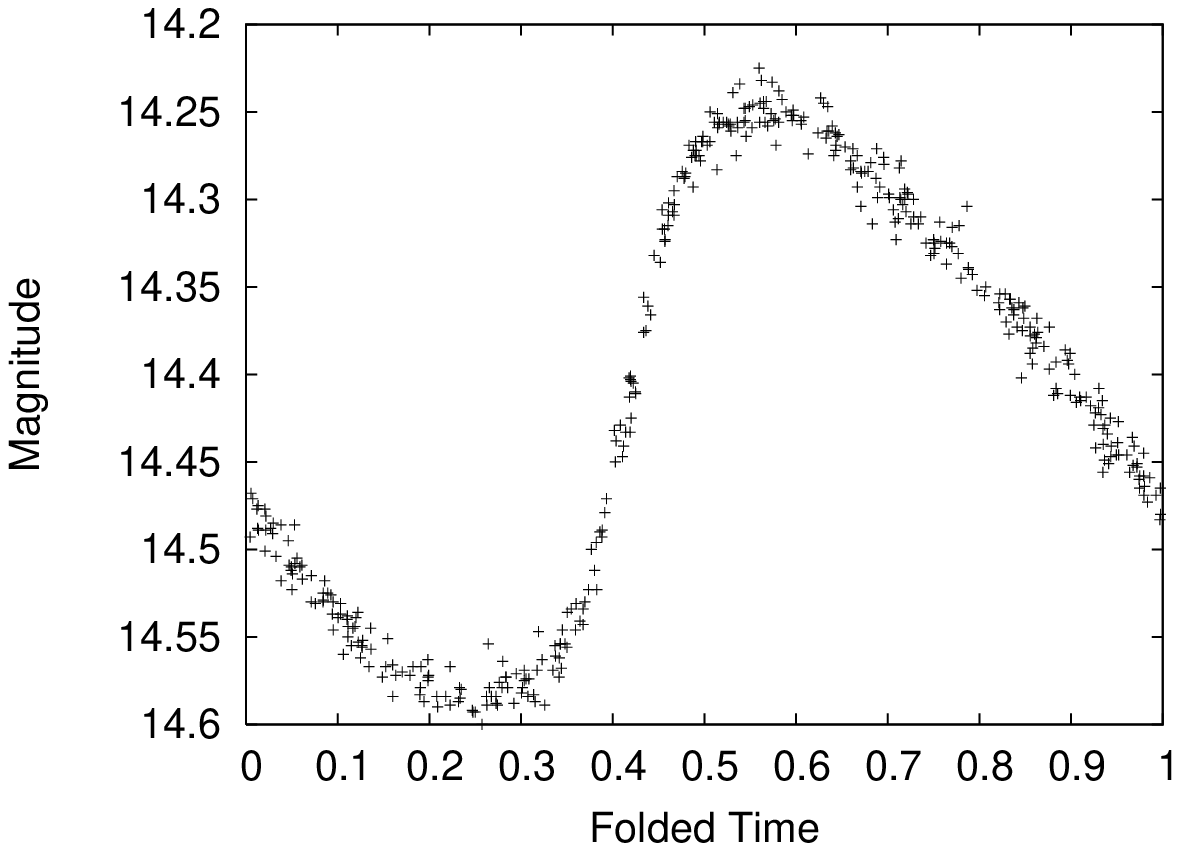} %
       \includegraphics[width=2in]{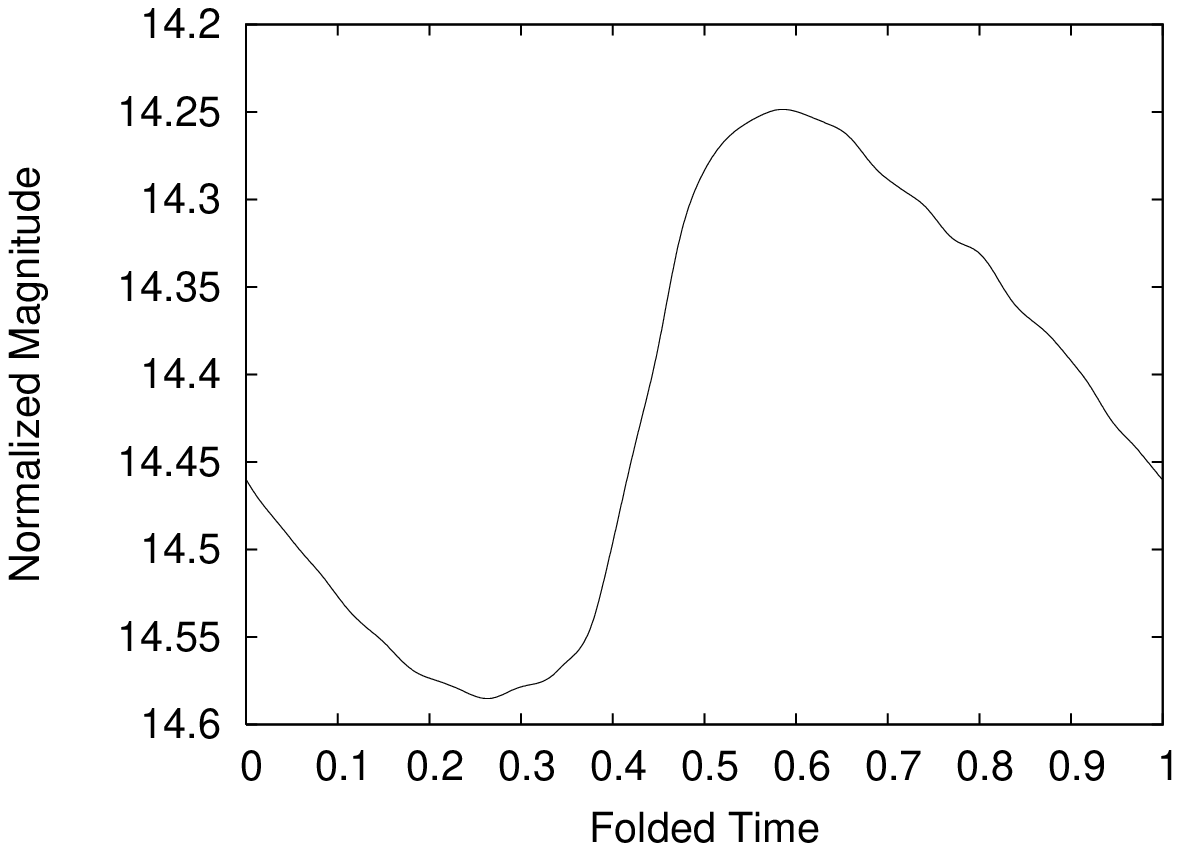} %
        } %
    }
    \mbox{
    \subfigure[Eclipsing Binary. OGLE052209.11-694441.9. Period 7.7 days.] {
           \label{fig:typical-eb}
       \includegraphics[width=2in]{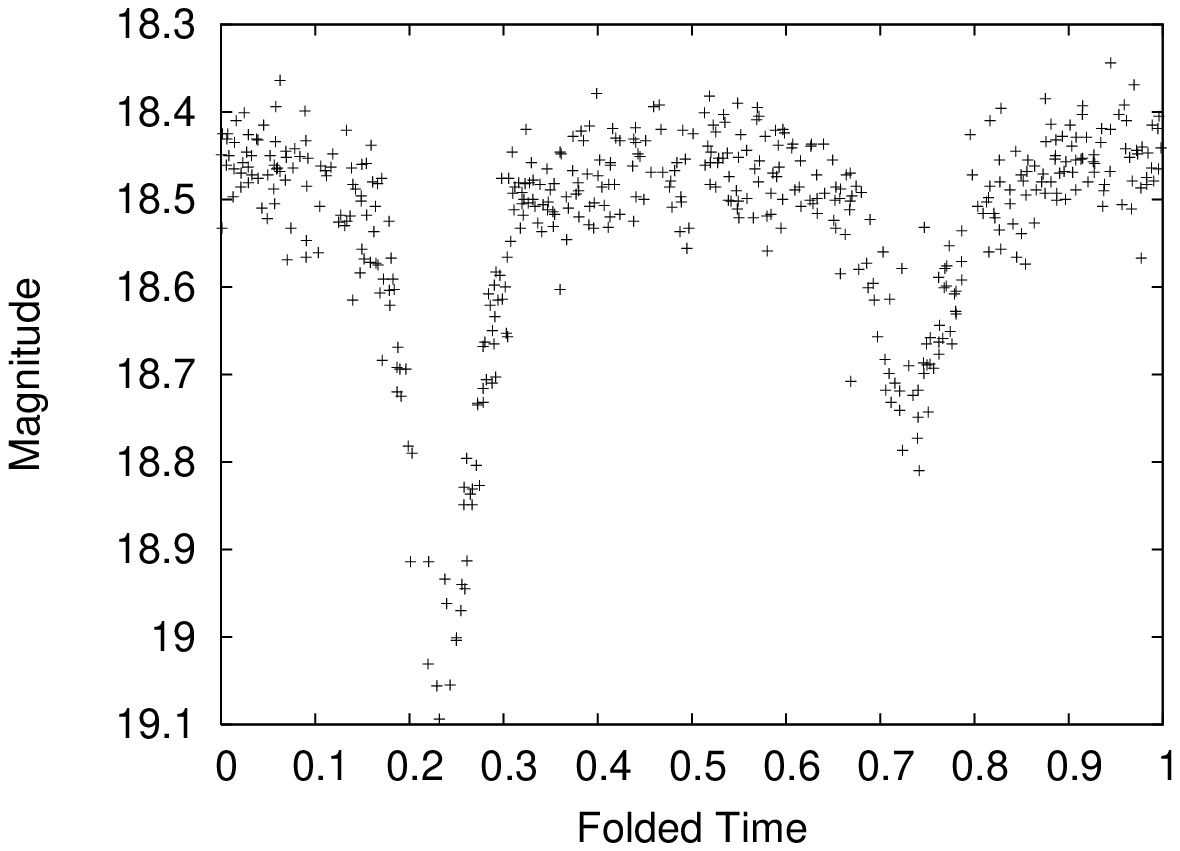} %
       \includegraphics[width=2in]{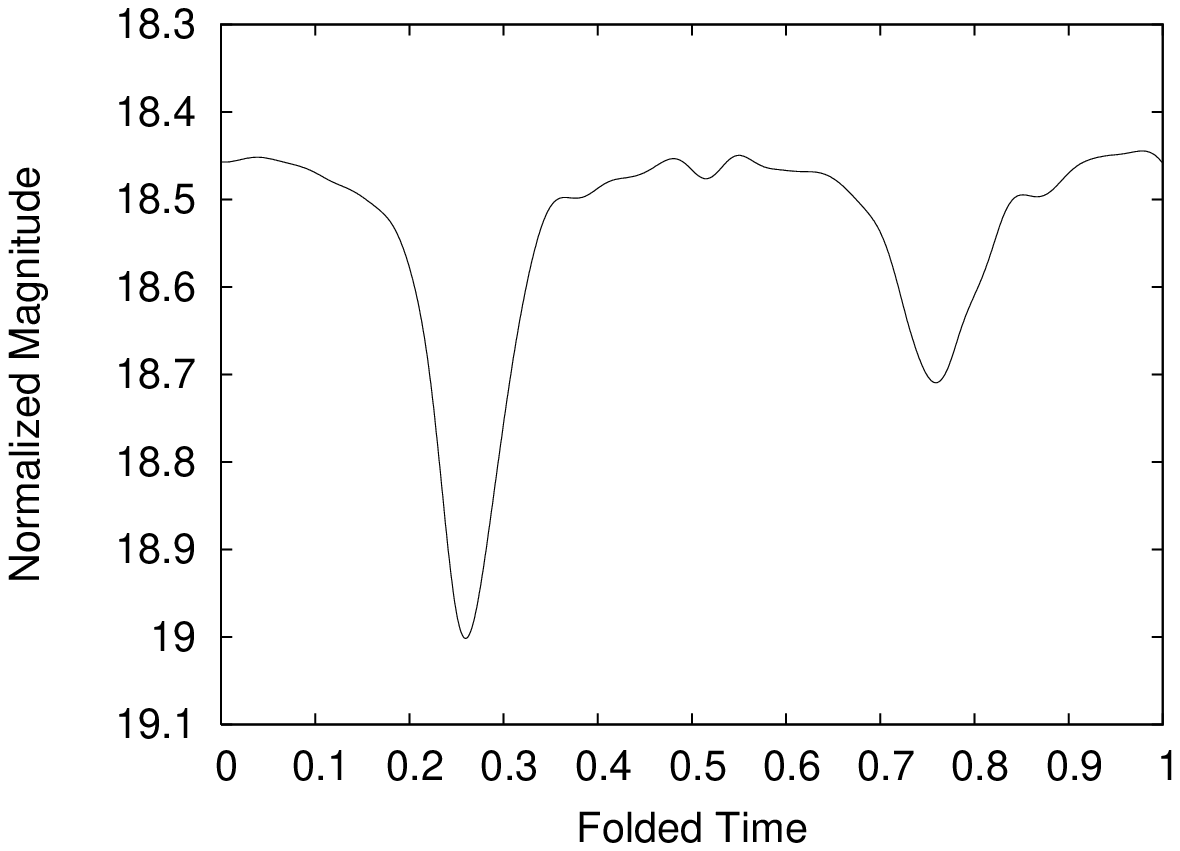} %
    } %
    } %
    \mbox{
    \subfigure[RR Lyrae. OGLE053520.04-703554.2. Period 0.34 days.] {
           \label{fig:typical-rrl}
       \includegraphics[width=2in]{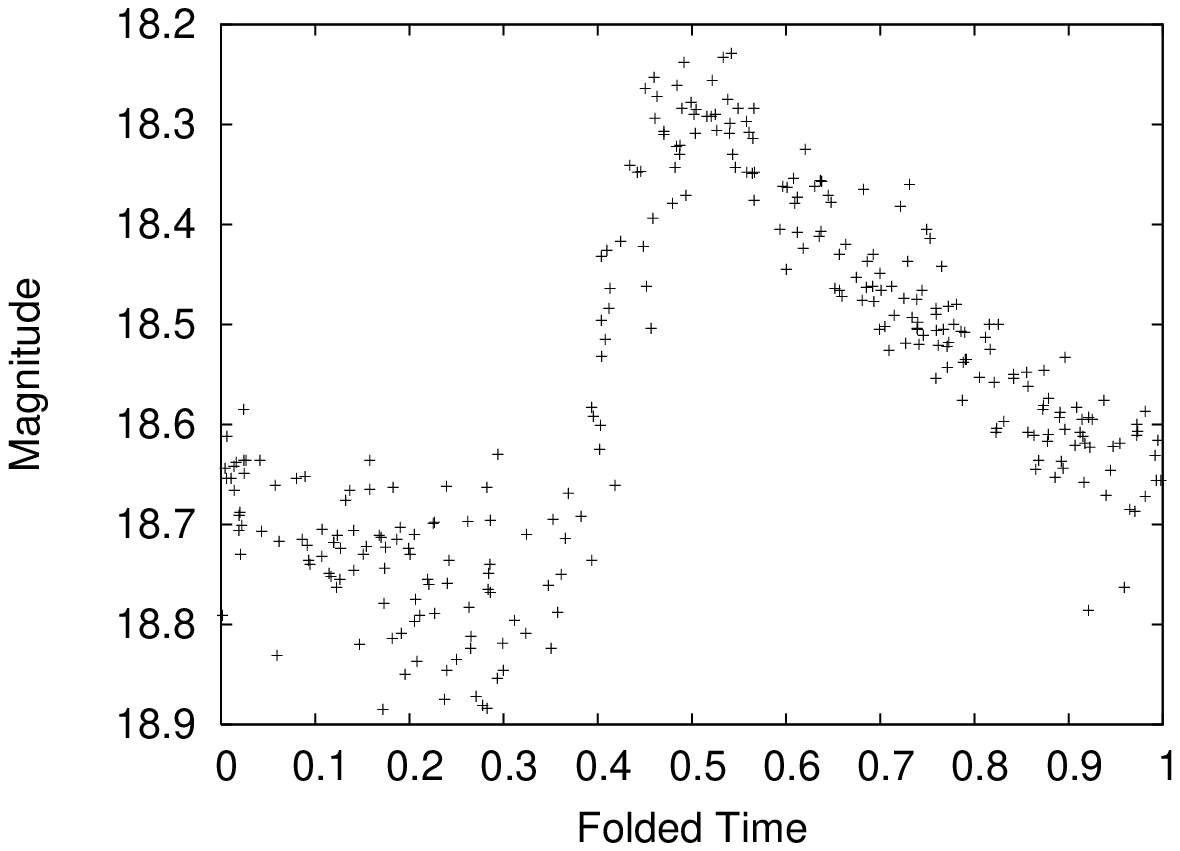} %
       \includegraphics[width=2in]{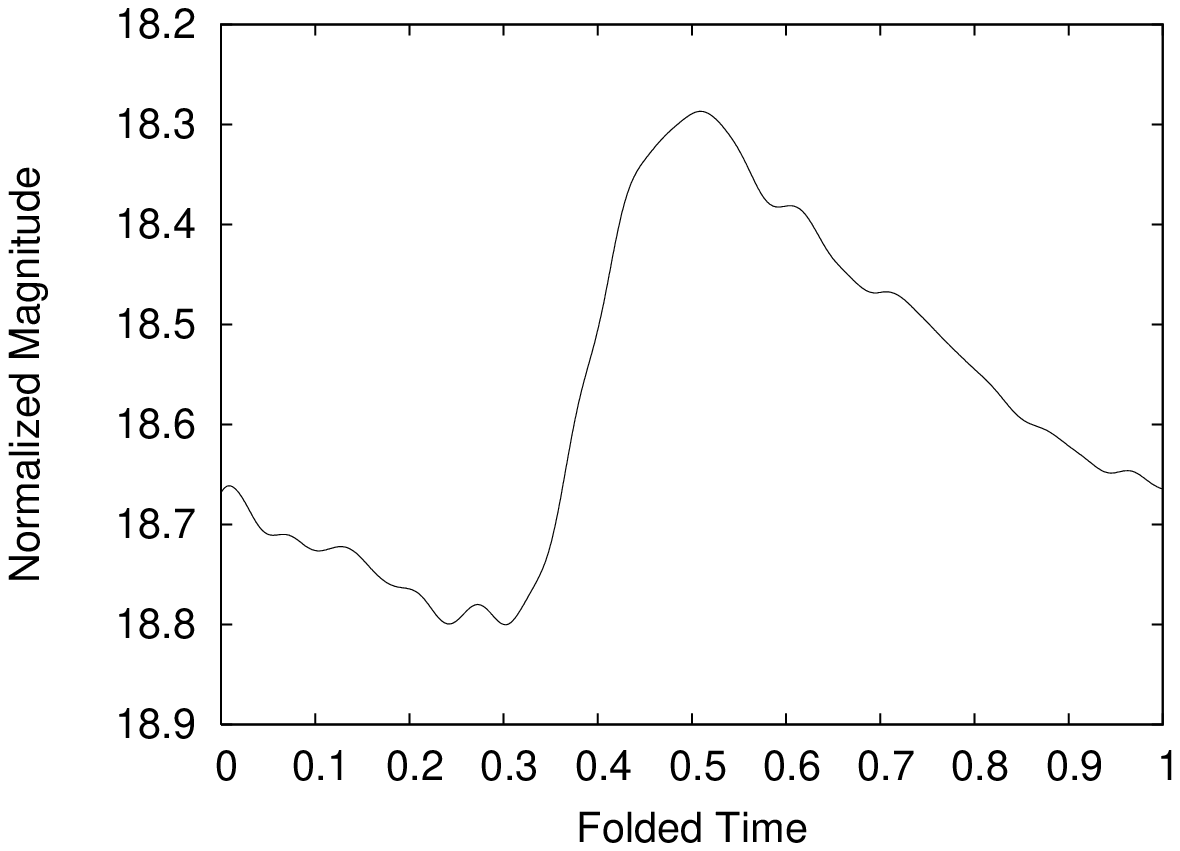} %
    } %
    } %
    \caption{ Examples of typical folded light-curves before (left)
    and after (right) pre-processing. For purposes of visual
    presentation, we align the maxima of each light-curve at
    approximately 0.25. This is also known as universal phasing, which
    is discussed in Section \ref{sec:challenge}.}
    \label{fig:typical}
\end{figure*}

Our research applies anomaly detection to photometric time series
data, called {\em light-curve} data.  Our specific application is to
find anomalies in sets of light-curves of periodic variable
stars. Most stars, like our own Sun, are of almost constant
luminosity, whereas variable stars undergo significant variations.
There are over 350,000 cataloged variable stars with more being
discovered.  The 2003 General Catalogue of Variable Stars \citep{samus2003}
lists known and suspected variable stars in our own galaxy, as well as
10,000 in other galaxies. For {\em periodic} variable stars, the
period of the star can be established. Common types of periodic
variable stars include Cepheid, Eclipsing Binaries and RR Lyrae, details of which can be found in
\citep{Petit87,Sterken96,Richter85}.  

The study of periodic variable stars is of great importance to
astronomy.  For example, the study of Cepheids yielded the
most valuable method for determining the Hubble constant, and the
study of binary stars enabled the discovery of a star's true
mass. Finding a new class or subclass of variable stars will be of
tremendous value.

Figure \ref{fig:typical} shows a typical light-curve from each star
class before and after we perform our pre-processing techniques
(described in Section \ref{sec:data}). The y-axis
measures the magnitude of brightness of the star.\footnote{Magnitude
refers to the logarithmic measure of the brightness of an object.} 
Magnitude is inversely proportional to the brightness of the
observation, thus, the y-axis is plotted with descending values.  The
x-axis measures folded time.  A folded light-curve is a light-curve
where all periods are mapped onto a single period, which is why there
may be multiple points on the y-axis for a single time point. We
describe light-curves and the process of folding in more detail in
Section \ref{sec:data}.

\begin{figure}[tb]
\begin{center}
\resizebox{0.75\textwidth}{!}{%
  \includegraphics[width=7cm]{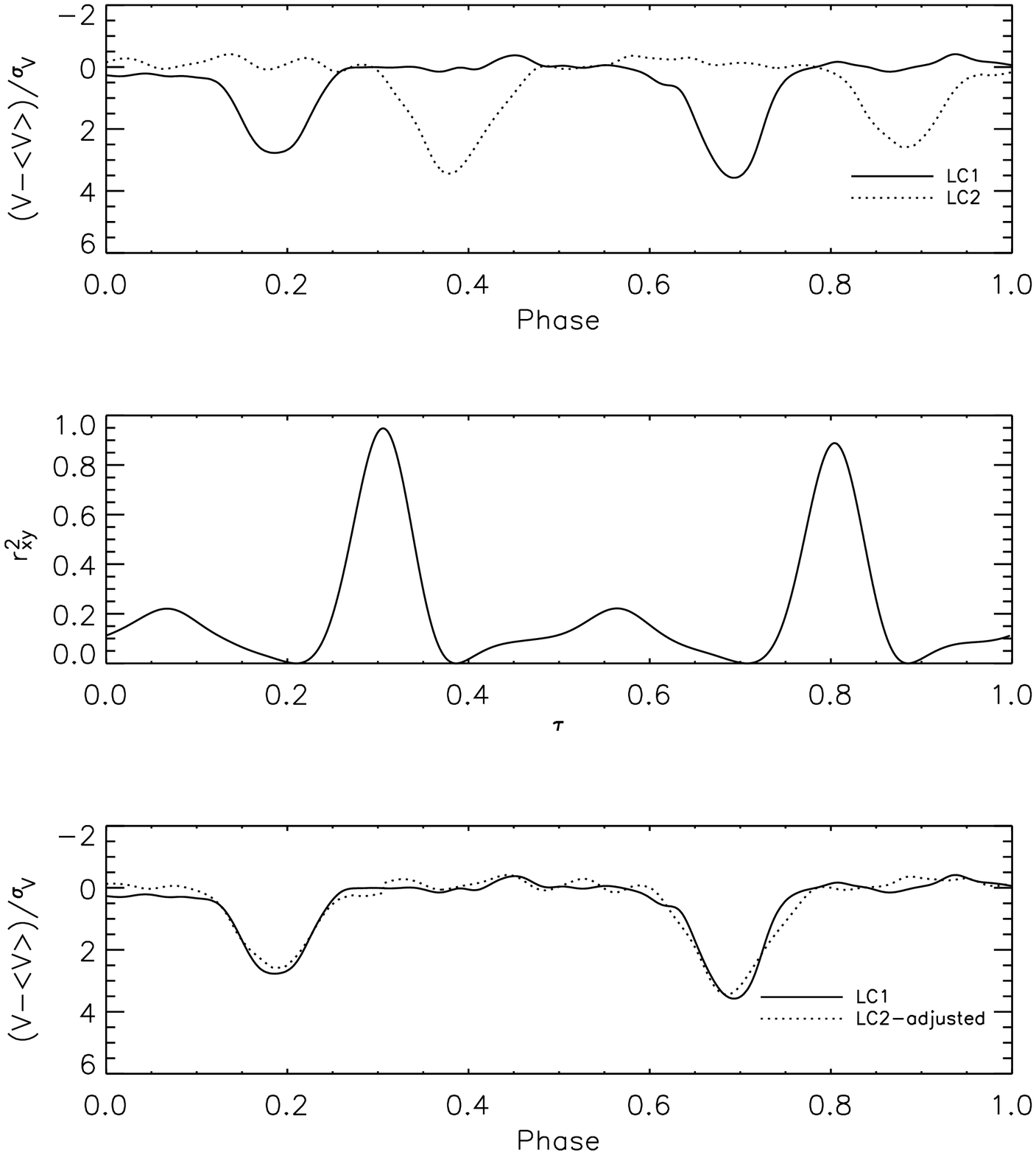}
}
\end{center}
\caption{The top panel shows two light-curves that are similar but
 whose phases are not synchronized.  The middle panel plots the square
 of the correlation as a function of the phase adjustment and shows that the
 global maximum occurs at a phase shift of approximately 0.3.  The
 bottom panel shows the light-curves after shifting the dotted curve
 to the right by this amount. }
\label{fig:phasing_example}
\end{figure}

Our research is motivated by the challenges inherent to performing
anomaly detection on large sets of periodic variable light-curves.
Several of these challenges are common to many time series data sets.
There are a large number of time-points in each light-curve (high
dimensionality), low signal-to-noise ratio, and voluminous amounts of
data. Indeed, new surveys, such as the Panoramic Survey Telescope and
Rapid Response System (Pan-STARRS), have the capacity to produce
light-curves for billions of stars \citep{panstarrs}. Any technique
developed for light-curves must scale to very large data sets.

A unique challenge of working with light-curve data is that the
periods of the light-curves are not synchronized because each is
generated by a different source (star).  To understand why phasing
poses such a challenge for anomaly detection in this domain, consider
Figure \ref{fig:phasing_example}, which illustrates how two similar
light-curves may appear dissimilar under a similarity measure like Euclidean distance if a phase adjustment is not
performed. The top panel shows two similar light-curves whose phases
are not synchronized. The middle panel shows the square of the
correlation plotted as a function of the phase adjustment.  The
maximum similarity occurs at a phase shift of approximately 0.3. The
bottom panel shows the two light-curves after the dotted light-curve
is shifted by this amount.  We define the optimal phase shift between
two light-curves as the shift that yields the maximum similarity
value.

This phasing problem presents a challenge to both general anomaly
detection techniques, and those developed specifically for time
series.  A general anomaly detection method, even with a metric that
works for unphased data, may not work out of the box.  With regard to
time series anomaly detection techniques, our task of finding
anomalies in $n$ distinct time series differs from most work which
assumes a single contiguous time series (not necessarily periodic) in
which anomalous sub-regions are sought. 

PCAD is our solution to the problem of anomaly detection on large sets
of unsynchronized periodic time series. The heart of PCAD is a modified k-means clustering algorithm, called Phased K-means (Pk-means), that runs on a sampling of the data.
Pk-means differs from k-means in that it re-phases each time series
prior to similarity calculation and updates centroids from these
rephased curves.  Because Pk-means is a modification of k-means, we
provide a proof that Pk-means does not break k-means's convergence
properties.

The Pk-means subroutine runs offline on a sampling of the data.  The
use of sampling enables PCAD to scale to large data sets.  The online
portion of PCAD is the calculation of the anomaly score for each time
series from the set of centroids produced offline by Pk-means.  This
operation is linear in the size of the data set.

Another advantage of PCAD is its flexibility to discover two types
of anomalies: local and global.  We define the terms local and global
anomaly and provide scoring methods for both.  Once each time series
is assigned an anomaly score, PCAD ranks the time series accordingly
and outputs the top $m$ for review.  To our knowledge, PCAD is the
only anomaly detection method developed specifically for
unsynchronized time series data that can output both global and local
outliers.

Our paper presents empirical evidence on four data sets that PCAD
effectively finds known anomalies and produces a better ranking of
anomalies when compared to naive solutions and other state-of-the-art
anomaly detection methods for time series.  We discuss the effect of
sample size and the parameter $k$ (used by Pk-means) on the anomaly
detection results, and show experimental results on light-curve data
with an unknown number of anomalies.  Our paper concludes with an
astrophysicists's discussion of the significance of the anomalies
found by PCAD.

\section{Related Work}
\label{sec:related-work}

PCAD is specifically developed to solve the problem of anomaly
detection on unsynchronized periodic time series data.  Because there
are few other methods with this purpose, we organize our related work
into three categories.  First, we briefly review general anomaly
detection methods for local and global outliers.  We then provide an
 in-depth review of anomaly detection methods for time series
data.  We conclude with a discussion of time series methods designed for unsynchronized data.

\subsection{Anomaly Detection}
\label{sec:general-anomaly}

Anomaly, or outlier, detection is a research area in both statistics
and data mining.  Statistical approaches focus on finding outliers
with respect to a particular statistical model or distribution, and
employ hypothesis testing for the discovery of outliers
\citep{hawkins,barnettlewis}.  Knorr and Ng define {\em distance-based
outliers} \citep{knorr98algorithms} as a point that is greater than
distance $D$ away from at least $p$ percent of the data set.  The
authors show that one can choose $D$ and $p$ to be
compatible with statistical definitions of outliers.  An advantage of the distance-based outlier definition is that it sidesteps the need
to assume a parametric form for the data. However, one must know or
search for the appropriate values of $D$ and $p$.  Other variants of
the distance-based outlier definition include the distance of a point
to its $k$-th nearest neighor \citep{ramaswamy00efficient}, or the sum
of the distances to its $k$-nearest neighbors \citep{angiulli02fast}.

The methods presented by Knorr and Ng include a simple brute-force
comparison of examples that avoids the use of indexing structures and
is $O(n^2)$, and a faster cell-based method that is $O(n)$ but
exponential in the number of features.  Subsequent approaches
improve on computational efficiency or scale to high-dimensional
data sets. Ramaswamy et. al's method improves efficiency by
partitioning the data into disjoint subsets and pruning calculations
among subsets containing no outliers \citep{ramaswamy00efficient}.  Bay
and Schwabacher employ pruning and randomization to achieve
near-linear time performance \citep{bay03mining}. Wu and Jermaine find
each example's $k$-nearest neighbors with respect to a random sample,
rather than the entire data set \citep{wu06outlier}.

The distance-based outlier definition does not correctly identify
outliers in mixed variance data.  Breunig et al. introduced the local
outlier factor (LOF) as a solution \citep{breunig00LOF}.  Rather than
setting a hard threshold $D$, LOF considers the density of an
example's neighborhood to calculate an outlier score.  Thus,
density-based outlier detection algorithms are able to discover {\em
local} outliers, or points that are anomalous with respect to their
nearest neighbors or assigned cluster, as opposed to {\em global}
outliers, points that are anomalous with respect to the entire data
set.  Other density-based outlier detection research improves the
computational efficiency of LOF by summarizing the data
\citep{jin01mining} and indexing \citep{ren04rdf}.

Both the distance- and density-based outlier definitions do not work
well in high dimensions due to the curse of dimensionality.  Solutions
to outlier detection on high dimensional data include feature
transformation \citep{angiulli02fast,yu04findout}, subspace projections
\citep{aggarwal01outlier,lazarevic05feature} and sampling
\citep{kollios03efficient,wu06outlier}.

In theory, the methods above can generalize to any type of data,
including time series data, if one were able to apply an appropriate
distance metric.  In Section \ref{sec:challenge}, we discuss why these
methods may not be appropriate for large sets of unsynchronized time
series data.  Moreover, all of the methods described above output either
distance-based global anomalies or density-based local anomalies.  In
contrast, PCAD has the flexibility to find both distance-based local
or global anomalies

\subsection{Time Series Anomaly Detection}

Methods for time series anomaly detection either operate on a single
time series or a time series database.  The goal of anomaly detection
on a single time series is to find an anomalous subregion.  The goal
of anomaly detection on a time series database is to find an anomalous
example. In some cases, a single time series is converted to a time
series database through the use of a sliding window.

Many time series methods supply their own definitions of anomaly,
rather than use the distance- and density-based definitions above.
Some define anomalies with respect to a {\em reference data set} of
``normal'' time series. The anomalies discovered in this context are
global anomalies. Other techniques use only the input data set, which contains
both normal examples and the unknown anomalies.  Depending on the amount of data used to compute their models, these techniques discover either global or local anomalies. For each technique,
we comment on whether the method makes use of a reference set.  For
techniques using only the input data, we comment on whether global or
local anomalies are sought.

In order to find an anomalous subregion within a single time series,
one may slide a window across the data either incrementally
\citep{dasgupta96novelty,Keogh-KDD-02,Ma-KDD-03,wei05assumption} or in
discrete steps according to a known period
\citep{yang01infominer,yang04mining}.  Dasgupta and Forrest
\citep{dasgupta96novelty}, and Keogh et al.'s \citep{Keogh-KDD-02}
methods use sliding windows and discretization on a single continuous
time series, and define anomalies with respect to a reference time
series.  They convert both their reference and input time series into
two time series databases by sliding a window of user-specified length
and discretizing the values in the window into a string.  In Forrest
and Dasgupta's work, a set of detectors is trained on the reference
set such that no detector will match any member of that reference set.
The detectors are then applied to the input set.  Any example that
signals an alarm is an anomaly. Keogh et al. encodes the set of
discretized strings from both the reference and input sets in suffix
trees.  The surprise score of each string in the input set is
calculated by comparing its number of occurrences in the input set
versus its expected number of occurrences.  These expectations are
calculated using a Markov model built from the reference set.  If the
actual number of occurrences differs from the expected number of
occurrences by more than a user-specified threshold, the string is flagged as
an anomaly.

Wei et al.'s method finds local anomalies without the aid of a
reference set \citep{wei05assumption}. The method discretizes the time
series using the SAX algorithm \citep{lin03symbolic} and then slides
two windows, called a ``lead'' and ``lag'' window, both of
user-specified length, across the data.  The lead window contains the
sequence to be labeled as normal or anomalous.  The lag window data
acts a reference set.  The lead window's anomaly score is the distance
between the lead and lag windows, where the distance metric calculates
the difference in frequency between all possible SAX subwords of a
given size. A drawback of this method is that a small cluster of
anomalies in the lag window might cause an anomalous pattern to be flagged as normal.

Discretization is often used to reduce the computation time of
operations on the sliding window. In the three methods described
above, comparisons are done in a discretized space.  A criticism of
this approach is that discretization may hurt the anomaly detection
process because too few symbols may overly smooth the data,
obfuscating some anomalies.  The degree of smoothing at which this
happens is dataset specific and remains an open area of research.

Ma and Perkins's method, also for a single continuous time series,
uses a sliding window of user-specified length but does not employ
discretization \citep{Ma-KDD-03}. Rather, it uses support vector
regression to model all previously seen subsequences.  For an
incoming time point, the sliding window shifts forward one point and
the newest input subsequence is evaluated with respect to the current
model.  If the input sequence differs from the model by more than a
user-specified tolerance, the sequence is anomalous.  The model is
continually retrained with each incoming time point.  Because the
computation time of the method increases with each incoming data
point, the authors propose dropping older samples in order to keep the
size of the training set constant.  A drawback of their approach is
that it requires the user to specify six parameters, including sliding
window size and kernel function.  The authors acknowledge having no
guidance for finding the optimal set of parameter values and declare
this to be an area of future work.

Unlike methods that seek anomalies of a pre-specified length, Shahabi
et al.'s method looks for anomalies at varying levels of granularity
(i.e., day, month, year) \citep{shahabi00tsatree}.  The authors develop
a tree structure called TSA-Tree that contains pre-computed trend and
anomaly information in each node.  The tree grows to depth $k$ where
granularity decreases as the tree-depth increases.  The authors use
feature extraction (wavelet filters) to capture trend and surprise
information at each granularity. However, the types of surprises are
limited to ``sudden changes'' in the data that are captured by ``local
maximums''.

Yang et al.'s InfoMiner technique detects ``surprising'' patterns on
periodic event sequence data \citep{yang01infominer,yang04mining}.
Thus, the data are already discretized, and the known period allows
the authors to treat a single continuous time series as a set of
smaller one-period time series.  InfoMiner detects global anomalies by
calculating information gain on each sequence, and labeling that
sequence as anomalous if its information gain exceeds a certain
threshold. Because this technique is for event sequence data, the use
of this program on real-valued time series data would require
discretization of the data set, and comparisons in discretized space.

The next two methods are developed for time series databases.
Salvador et al.'s method creates a finite automaton trained on a
reference set of time series data \citep{salvador04learning}.  An input
time series is anomalous if it does not conform to the model, thus
ending at an anomalous state. The state transition logic for the
automaton is determined by a three step process.  First, each
reference time series is segmented using a novel clustering algorithm
called Gecko that determines the approriate number of segments.
Second, the slope is extracted from each segment and then mapped to
rules that are, in the third step, converted into state-transition
logic. Time does not figure into the features that are used to map
segments to states.  This meets the specific needs of their
application (faulty valve detection), but would not work for a domain
in which the relationship of the signal to time must be preserved.

Jagadish et al. find ``deviants'' in a time series database by
modeling the time series with a set of minimum-bounding rectangles
or ``histograms'' \citep{Jagadish-VLDB-99}.  Dynamic
programming is used to find the optimal number of rectangles and
rectangle-width.  The $k$ most deviant time points are first
identified.  These are points whose removal best improves the fit
between the histogram and the original time series.  After identifying
deviant points, the method then identifies anomalous subregions as
those having a number of deviants that exceeds a user threshold.
Jagadish et al.'s technique, which finds global anomalies, is
quadratic in the length of the time series and therefore too slow for
light-curve and other voluminuous time series data.

\subsection{Time Series Methods for Unphased Data}

The majority of time series methods that tackle the problem of unsynchronized
data have the goal of time series clustering
\citep{barjoseph2002new,chudova03translation,gaffney04joint}.  These
methods are applied to a diverse set of applications including the tracking
of cyclone trajectories \citep{chudova03translation}, gene expression
\citep{gaffney04joint,chudova03translation,barjoseph2002new}, and
botany \citep{listgarten06bayesian}.  Clearly, unsynchronized time series data
sets exist in domains outside of astrophysics.

Of the methods cited above, only \citep{gaffney04joint} and
\citep{chudova03translation} integrate data synchronization into their
clustering methods.  Both use the EM algorithm
\citep{dempster77maximum} to estimate cluster membership as well as the amount
of horizontal shift that occurs in each time series.  Gaffney and Smyth
develop a fully Bayesian generative model in ``curve space'', which means
they represent the curves using a regression mixture model.  Thus, the time
series need not be of uniform length. Their joint clustering-alignment model
also assumes a normal regression model for the cluster labels, and Gaussian
priors on the (hidden) transformation variables, where transformations
include shifting and scaling on both the time axis and in measurement space.
Chudova et al. \citep{chudova03translation} develop a generative model that
assumes uniformly spaced time points, but also multi-dimensional curves.
Their model is formed in the time domain where the distribution of each curve
minus its offset in both time and measurement space is assumed to be
multivariate Gaussian with a diagonal covariance matrix.  The authors employ
conjugate priors on the model parameters, and use Gibbs sampling to obtain
point estimates of both parameters and hyperparameters.  Note that the goal
of both methods is to produce the best clustering model.  In contrast, PCAD
uses a much simpler clustering method (that makes no parametric assumptions)
for the purpose of anomaly detection.  Their results show that applying EM
to these Bayesian generative models produces good clustering results on
unsynchronized data.  However, for our application, these computationally intensive
methods are unnecessary for the purpose of anomaly detection.

We discuss in detail the methods that do anomaly detection on unsynchronized time series data.  Chan and Mahoney's technique creates a model of minimum-bounding
rectangles from a reference set of $n$ normal time series
\citep{chan05multiple,mahoney05trajectory}.  Before computing this
model, each time point in each time series is mapped to a
three-dimensional vector where the attributes are the smoothed time
point plus its smoothed first and second derivatives. A box model is
created for each reference time series using a greedy algorithm that
finds $x$ boxes that minimize the volume over the three-dimensional
feature space. Given the $n$ separate box models, the two closest
boxes across the different time series are repeatedly merged until
only $k$ remain.  An anomaly score is calculated for an input time
series by transforming the series into the three-dimensional feature
space and summing each point's distance to the nearest box in the
model.  Because this algorithm trains a model on multiple time series,
the authors also encounter the phasing issue, which they acknowledge with
the statement: ``two time series with corresponding behavior might
not begin and end at the same time in the time series.''  The authors
handle phasing by allowing any two boxes to merge, regardless of their
order in the original time series. It is our opinion that this
``order-independent'' algorithm may be too lenient, because it is
possible to completely rearrange a time-series with this method.  In
contrast, the phase adjustment done by PCAD shifts a time series
horizontally but otherwise respects the order of points. Also, Chan
and Mahoney's method requires a reference set containing only typical
examples from which a model of normality is calculated.  PCAD does not
use a reference set because a comprehensive set of typical examples is a
luxury our domain experts may not be able to provide.

Wei et al.'s rotation-invariant discord discovery technique, which we
refer to as RI-DISCORD, \citep{wei06saxually} is an anomaly detection
methods that discovers global anomalies or {\em discords} in shape
data. Discords, whose definition is reminiscent of the distance-based
outlier definition, are the $m$ most anomalous time series (or
subregions if the input is a single time series) with the farthest
nearest neighbors \citep{keogh05hotsax}. The definition can also be
extended to find the top $m$ examples with the farthest distance to
their $k$-nearest neighbors. RI-DISCORD takes a database of shape data
as input, each of which is converted to a time series.  In order to
ensure that different rotations of the same shape matches another, Wei
et al. use a rotation-invariant Euclidean distance measure that searches over
all possible rotations of the time series and outputs the minimum
distance calculated.

The $O(n^2)$ brute-force version RI-DISCORD is identical to the brute-force version of Keogh
et. al's discord detection method called HOT SAX \citep{keogh05hotsax}.
The only difference is the use of the rotation-invariant distance measure in RI-DISCORD.  Both have faster versions that
use heuristics to order the inner and outer loops of the search in a
way that ensures maximal pruning of distance calculations, but they
differ in the heuristics used.  A drawback of both methods is that
they do not identify small clusters of similar anomalies. The
RI-DISCORD method attempts to resolve this by extending the definition
of discords to be the top $m$ examples with the largest distance to
its $k$th nearest neighbor.  However, this requires one to know in
advance the value of $k$ that is inclusive of small anomalous
clusters.

Because RI-DISCORD is the only anomaly detection method that is
designed to handle unphased data without the use of a reference set,
we compare PCAD's performance to RI-DISCORD in Section
\ref{sec:results}.  Note, that RI-DISCORD is designed to find global
anomalies only.  It does not discover local anomalies.

\section{The Challenge of Unphased Data}
\label{sec:challenge}

One may reasonably ask whether any of the methods cited in Section
\ref{sec:related-work} can be adapted or applied to handle
unsynchronized time series data.  The answer is no for methods that
perform feature transformations that do not respect the time domain
\citep{aggarwal01outlier,lazarevic05feature}.  These methods would not
work for time series data in general.  For methods that rely on
spatial indexes \citep{ramaswamy00efficient}, one would need to
synchronize or {\em universally phase} the data before applying the
anomaly detection method.  We discuss universal phasing at the end of
this section.

The other way to adapt a general outlier detection method is to use
cross correlation, a distance metric for which there exists an
efficient method for computing the optimal phase shift that maximizes
the similarity between two time series.  For time series $x$ and $y$,
cross correlation is computed

\begin{equation}
\label{Equ:CrossCorrelation}
  r^2_{xy}(\tau) = \sum_{t=0}^{d-1} x(t) \, y(t-\tau) \: , 
\end{equation}

\noindent where $t$ is discrete time, $d$ is the number of time points
and $\tau$ is the phase shift.  A phase shift implies a rotation of the times series.  If $t-\tau < 0$, we wrap the time series around and let $y(t-\tau)$ equal $y(t-\tau+d)$. Finding
the max of $r^2_{xy}(\tau)$ requires finding the value of $\tau$ that
maximizes cross correlation. The running time of a brute force search for the optimal
$\tau$ is $O(d^2)$.  Fortunately, we can use the convolution theorem
to find the same value in $O(d\log{d})$.  According to the convolution
theorem, cross correlation can be written as

\begin{displaymath}
\label{Equ:CorrelationFreqDomain}
  r^2_{xy}(\tau)
          =    {\cal F}^{-1}  \left[    {\cal
X}(\nu)  \, \bar{{\cal Y}}(\nu)    \right]  (\tau)
\end{displaymath}

\noindent where  ${\cal X}(\nu)$ is the Fourier transform of $x(t)$ and
$\bar{{\cal Y}}(\nu)$ is the complex conjugate of the Fourier
transform of $y(t)$. One finds the maximum correlation
by finding the maximum of the inverse Fourier transform of the
product of the Fourier transforms of the two time series.  For fast
Fourier transforms (FFT), each Fourier transform requires $2d
\log{d}$ operations for any value $d$. Thus, the
calculation of cross correlation of a pair of time series is
$O(d\log{d})$. 

Simply substituting cross correlation maximized over all possible
phase shifts, or {\em maximized cross correlation}, as the distance
metric into a general outlier detection method may not yield desirable
results for unsynchronized time series data.  First, the use of
maximized cross correlation increases the computational expense of
distance calculations from $O(d)$ to $O(d \log{d})$, a non-trivial
expense if repeated $k$-nearest neighbor calculations
 are performed on high-dimensional data \citep{bay03mining,keogh05hotsax}.  This cost
might be absorbed by methods running on data sets where $n >> d$.
However, a method that aggregates the results of distance calculations
(e.g., averaging) \citep{jin01mining} on a subset of the data must
contend with the set of unique phase shifts for every pair of time
series in that group, and possibly rephase individual time series before performing those
calculations.  One may need to redesign the method to intelligently
handle these issues.

Universal phasing is the global synchronization of the data set, after which, an
out-of-the-box anomaly detection method can be applied.  The simplest
universal phasing algorithm aligns a set of times series by
setting the maximum (or minimum) of each time series to time $t$.
Protopapas et al. developed a more robust version of this method that
accounts for noise \citep{protopapas05b}.  Rather
than phase each time series to its maximum, the method sorts the values of the time series and isolates the highest and lowest ten percent.  Those values are then separated into two clusters.
Time $t$ is set to the mean value of the cluster containing the higher (or lower)
values. For full details on how universal phasing is performed, please
refer to \citep{protopapas05b}.  

The drawback of universal phasing is that it does not work for all
data sets.  Specifically, it will not work if the data set contains
time series with multiple maxima (and minima) of the same height (give
or take noise).  Under those circumstances, the phasing algorithm will
arbitrarily choose one peak and set that to time $t$.  Similarity
calculations will be incorrect if performed on time series that do not
universally phase well.

For some data sets, the general shape of the time series is known and
universal phasing is appropriate.  However, for large sets of time
series containing an unknown classes of objects, the use of universal
phasing poses a risk.  The imminent release of the Pan-STARRS project
underscores this urgency.  Billions of light-curves of unknown class
are forthcoming.  There is no guarantee that the shape of all objects
from this survey will meet the requirements for universal phasing.
Given the pending volume and diversity of this project, it is prudent
to use a method that makes no assumptions on the shape of the time
series, and whose computational expense already accounts for the cost of
phasing.

\section{PCAD}
\label{sec:PCAD}


%

PCAD, an acronym for Periodic Curve Anomaly Detection, is our anomaly
detection technique for sets of unsynchronized periodic time series.
PCAD does not use a reference data set of normal examples.  Instead,
it uses unsupervised learning on a sampling of the data to generate a
set of centroids that are representative of the data.  The local and
global anomaly scores for a time series is based on its similarity to
this set of centroids.

We implement a modified k-means algorithm as our unsupervised
technique.  Our modification solves the phasing issue by rephasing each time series to its closest centroid
before recalculating the centroid at every iteration.  Our algorithm, Pk-means, produces a
set of $k$ centroids that are used to calculate the local and global anomaly scores.
The end user may sort the time series data according to his desired anomaly score
and obtain a ranked set.  

We first review the k-means algorithm before describing the Pk-means algorithm in-depth.  We then discuss the computational expense of Pk-means and how sampling the data improves the overall efficiency of PCAD. We then discuss the final step of PCAD, outlier score calculation, and provide scoring functions for both local and global outliers.

\subsection{Review of k-means}

Given a set of $n$ data points, k-means begins by initializing the
cluster centers to a set of $k$ points randomly selected from the
data.  It then calculates the similarity between each data point and
each centroid, (re)assigns each point to its closest centroid, and then
calculates a new centroid for each cluster by averaging the points in
the cluster.  The algorithm repeats from the similarity calculation
step until a convergence criterion is met (e.g., clusters no longer
change).

The time complexity of k-means in $O(knrd)$ where $n$ is the number of
data points (or time series, in our case), $k$ is the number of
centroids, $r$ is the number of iterations until convergence and $d$
is the number of features, which in our case, is the number of points
in the time series.  In order to run k-means, either the user must
supply the value of parameter $k$, or use a model selection criteria
(e.g., the Akaike Information Criterion or the Bayesian Information
Criterion \citep{pelleg00xmeans}) to automatically determine an
appropriate value of $k$.

\begin{figure}[tb]
\begin{algorithmic}[1]
   \STATE Pk-means(Time Series Set: x[],  Number of centroids: $k$)
   \STATE Initialize centroids[]
   \WHILE{ ! AchievedConvergence }
   \STATE best\_centroids[] $\leftarrow$ CalcDistance(x[], centroids[])
   \STATE clusters[][] $\leftarrow$ AssembleClusters(x[], best\_centroids[])
   \STATE centroids[] $\leftarrow$ RecalcCentroids(clusters[][])
   \ENDWHILE
   \STATE return centroids[]
\end{algorithmic}
\caption{Algorithm for Pk-means.  Data structures followed by a [] denote list or array structures.}
\label{alg:modified-kmeans}
\end{figure}

\begin{figure}[tb]
\begin{algorithmic}[1]
\STATE CalcDistance(Time Series set: x[], Centroids: centroids[])
\FOR{$i=1$ to $n$}
   \STATE max\_corr[$i$] $\leftarrow$ 0 
   \FOR{$j=1$ to $k$}
      \STATE ($corr, phase$) $\leftarrow$ CalcCorrelationFFT(x[$i$],centroids[$j$])
      \IF{max\_corr[$i$] $<$ corr}
         \STATE max\_corr[$i$] $\leftarrow$ corr
         \STATE best\_center $\leftarrow j$
         \STATE best\_phase $\leftarrow$ phase
      \ENDIF
   \ENDFOR
   \STATE best\_centroids[$i$] $\leftarrow$ best\_center
   \STATE x[$i$] $\leftarrow$ UpdatePhase(lc[$i$], best\_phase)
\ENDFOR
\STATE return best\_centroids[]
\end{algorithmic}
\caption{Algorithm for the distance calculation subroutine of Pk-means.  Data structures followed by a [] denote list or array structures.}
\label{alg:calc-distance}
\end{figure}

\subsection{Pk-means}
\label{sec:pkmeans}

In Figure  \ref{alg:modified-kmeans} we show the pseudo-code for Pk-means. The
initialization and cluster assembly subtasks
remain unchanged from k-means.  Convergence is achieved
when cluster composition does not change upon successive iterations.
We discuss convergence in detail in Section \ref{sec:convergence}.

\begin{figure}[tbp]
    \centering
    \mbox{
    \subfigure[Cepheid] {
           \label{fig:centroids-ceph}
       \includegraphics[width=2in]{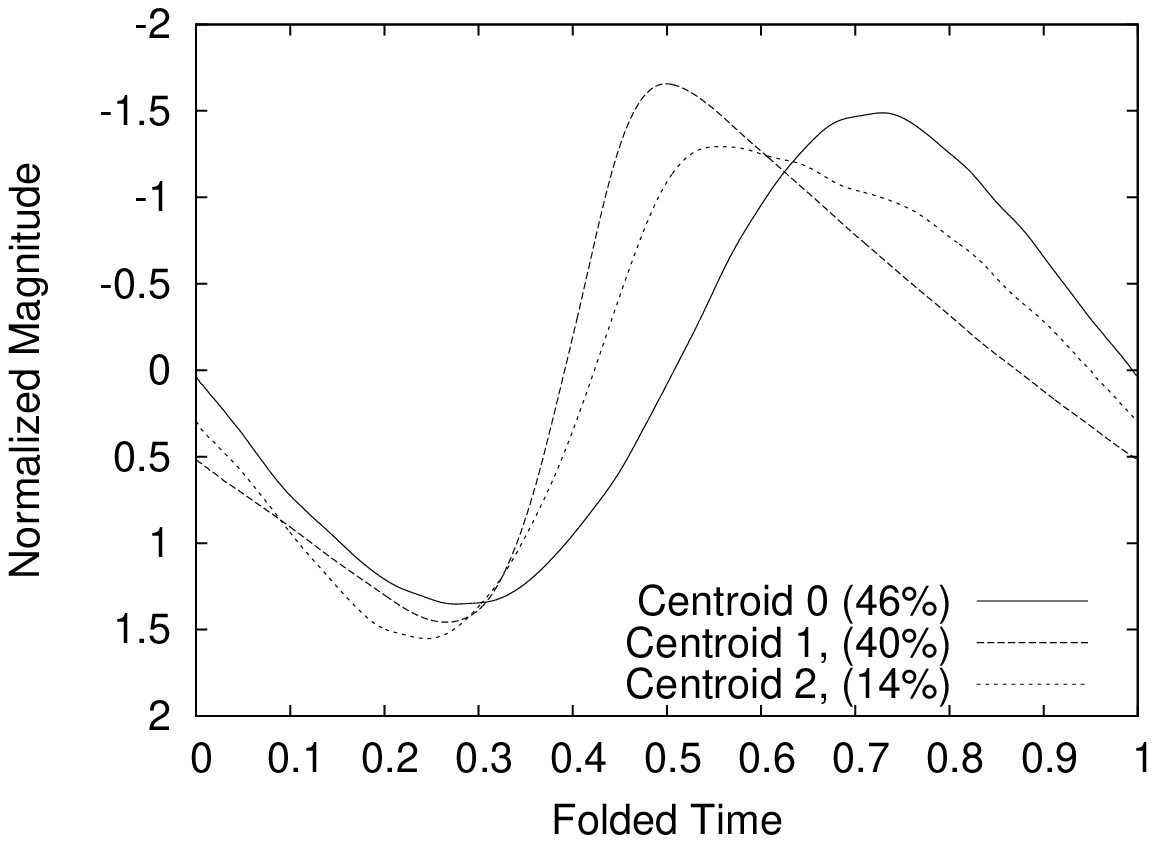} %
        } %
    \subfigure[Eclipsing Binary] {
           \label{fig:centroids-eb}
       \includegraphics[width=2in]{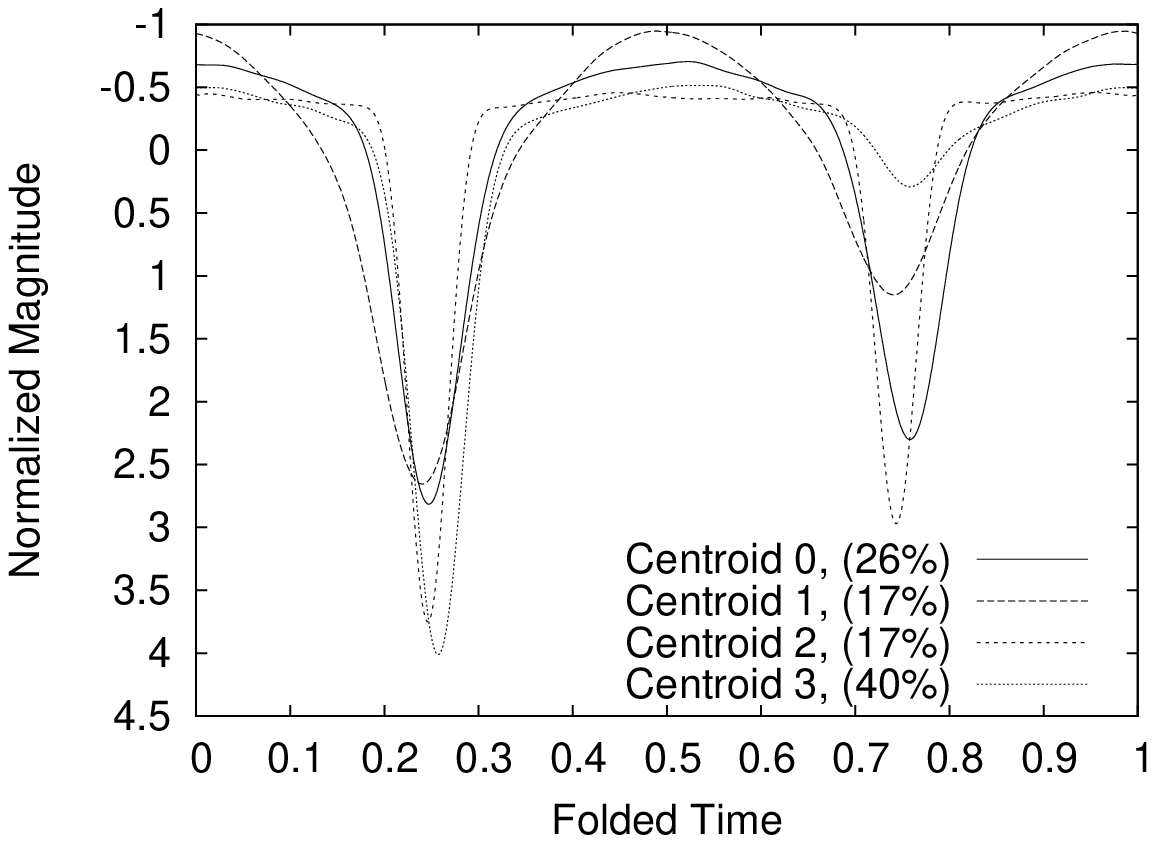} %
    } %
    }
    \subfigure[RR Lyrae] {
           \label{fig:centroids-rrl}
       \includegraphics[width=2in]{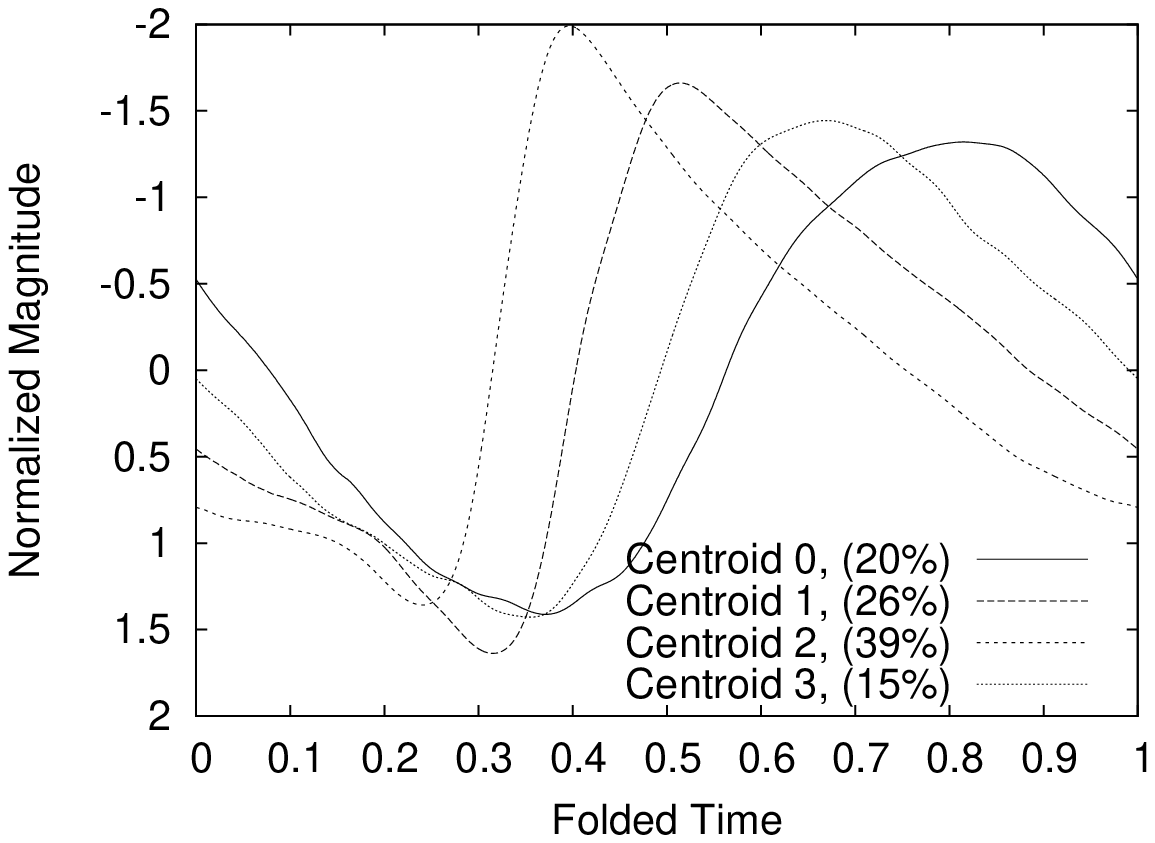} %
    } %
        \caption{Centroid output of Pk-means.  The number in
        parentheses is the percentage of the data assigned to that
        cluster.  The parameter $k$ equals 3, 4 and 4 in these figures
        respectively.  In our experiments, $k$ is chosen
        using the Bayesian Information Criterion (BIC).
        }
    \label{fig:centroids}

\end{figure}

Figure \ref{alg:calc-distance} shows the distance calculation
subroutine used by Pk-means.  For each time series $x[i]$, we
calculate maximized cross correlation between it and each of the $k$
centroids.  We determine which centroid is most similar to $x[i]$
and store the phase adjustment required to produce the maximum correlation.
The centroid number is stored in the array $best\_centroids[i]$
and used for cluster assignment in the next step of Pk-means.
However, before exiting the distance calculation subroutine, $x[i]$
is rephased according to the phase adjustment needed to maximize its
correlation to the centroid.

This last step is critical; upon each iteration, Pk-means adjusts the
phase of each time series. In k-means, the data points never change;
centroid calculation is only affected by cluster membership.  In
Pk-means, each centroid is determined by its cluster membership and
the phase adjustments of the time series in its cluster.  Thus,
Pk-means is not merely k-means using cross correlation as a distance
metric.  Figure \ref{fig:centroids} shows examples of centroid
light-curves output by Pk-means.

As with k-means, the parameter $k$ must be specified in advance. 
We search across multiple values of $k$ and select the optimal $k$
using the Bayesian Information Criterion (BIC). BIC is calculated
according to the method outlined in \citep{pelleg00xmeans}.  The set of
centroids produced from this run of Pk-means is used to calculate the
anomaly scores for all light-curves in the data set.

\subsection{Sampling and Computation Time}

The time complexity of Pk-means in $O(knrd\log{d})$ where $n$ is the
number of time series, $k$ is the number of centroids, $r$ is the
number of iterations until convergence, $d$ is the number of points in
a single time series, and $d\log{d}$ is the cost of computing cross
correlation using a FFT.  Because $d$ and $k$ are independent of $n$,
we can reduce the time complexity to $O(nr)$.

Time series methods must be $O(n)$ in order to scale to large data
sets.  At $O(nr)$, Pk-means does not meet this requirement.  However,
recall that the goal of Pk-means is to produce a set of centroids that
is representative of the data.  Given a large data set, it may be
possible to generate this representative set from a random sampling of
size $s$ where $s << n$.  This reduces the time complexity of Pk-means
with respect to the overall data set size.  Additionally, we can
perform Pk-means (as well as the associated search for the optimal
$k$) offline.  In Section \ref{sec:ss-effect}, we present empirical
results that support the intuition that sampling the data to find the
centroids has little impact on the rankings of the anomalies.

The online portion of PCAD is simply the anomaly score calculation.  A
local and global anomaly score can be simultaneously calculated for
each time series in the data set.  The anomaly score calculation uses
the FFT method of calculating maximized cross correlation between each
time series and the centroids output by Pk-means.  Thus, the
computation time of the anomaly score calculation is $O(knd\log{d})$
which reduces to $O(n)$.  Thus, the online portion of PCAD is linear
in the size of the data set.

\subsection{Local and Global Anomaly Scores}

We calculate the anomaly score for a time series in relation to the
$k$ centroids returned by Pk-means.  If one seeks global anomalies,
then the anomaly score for time series $x[i]$ or $x_i$ is calculated:

\begin{equation}
\label{eq:anomaly-score-global}
score(x_i) = \sum_{j=1}^k \frac{|c_j|}{n}r^2_{x_i,c_j}
\end{equation}

\noindent where $c_j$ is a centroid, $|c_j|$ is the number of time
series closest to $c_j$, and $n$ is the size of the data set.  The
lower the score, the more anomalous the time series is with respect to
the entire data set.  We use a weighted average to offset the
influence of relatively small clusters.  


To find local anomalies, the anomaly score is the distance of a time
series to its closest centroid.  This is computed as:

\begin{equation}
\label{eq:anomaly-score-local}
score(x) = \min_{j} r^2_{x,c_j}
\end{equation}

After an anomaly score is calculated for each time series,
the time series are sorted in order of ascending scores.  For global anomalies, the top $m$
time series (where $m$ is specified by the user) are returned as
the list of anomalies.  For local anomalies, we isolate the top $m$ anomalies per cluster.


\section{Pk-means Convergence}
\label{sec:convergence}

In this section we prove that Pk-means, and its use of maximized cross
correlation, does not break the convergence properties of k-means. We
first review the proof of convergence for k-means and then prove the
convergence of Pk-means.

\subsection{Review of Proof of Convergence of K-means}
\label{sec:convergence-kmeans}

The following proof of k-means' convergence has been adapted from
\citep{bottou95convergence}.  Please note that we have slightly
changed the notation. 

If we review the steps of k-means, we see that each iteration begins
with a set of centroids $W$, which are either the $k$ centroids
k-means is initialized with, or the centroids calculated during the
previous iteration.  Within each iteration of k-means, a new set of
clusters $C$ is assembled based on each data point's proximity to
the centroids $W$.  From $C$, a new set of centroids $W'$
is calculated, from which a new cluster assignment is made $C'$.
Thus, k-means progresses from 
\begin{equation}
W \rightarrow C \rightarrow W' \rightarrow C' \nonumber
\end{equation}

We define the quantization error $E(W,C)$ in relation to $W$ and $C$ as follows:

\begin{eqnarray}
 E(W,C) & = & \sum_i \frac{1}{2} \min_{w \in W} (x_i - w)^2 \nonumber \\
      & = & \sum_i \frac{1}{2}(x_i - w_{c(i)})^2  \label{Equ:E}
\end{eqnarray}

\noindent where $x_i$ is a data
instance, $w$ is the centroid in $W$, and $c(i)$ is the
cluster to which $x_i$ is assigned.  

K-means converges if, at each iteration, the quantization error is
non-increasing. To show that the quantization error is non-increasing, we must prove
$E(W',C') \le E(W,C)$, or

\begin{eqnarray}
\sum_i \frac{1}{2}(x_i - w'_{c'(i)})^2 & \leq & \sum_i \frac{1}{2}(x_i - w_{c(i)})^2  \nonumber
\end{eqnarray}

To show that the error decreases between $E(W,C)$ and $E(W',C')$, we
demonstrate that $E(W',C') \le E(W',C) \le E(W,C)$. This decreasing
progression corresponds to the steps of the k-means algorithm.  The
term $E(W',C)$ corresponds to the decrease in error that results after
calculating the new centroids, while maintaining the old cluster
assignments, where

\begin{eqnarray} 
\label{Equ:Q}
E(W',C) & = & \sum_i \frac{1}{2}(x_i - w'_{c(i)})^2  \nonumber
\end{eqnarray}

The term $E(W',C')$ results after the cluster assignments are updated
so that each point is assigned to the cluster of its closest centroid
under $W'$.

In the k-means algorithm, the new centroids $W'$ are calculated by
averaging the points in each of the clusters in $C$.  This definition
of $W'$ is the one that minimizes $E(W',C)$. We find the value for $w'
\in W'$ by computing ${\delta}E/{\delta}w'$, setting the resulting
expression equal to 0, and solving for $w'$:

\begin{eqnarray}
w' & = & \frac{1}{|c|} \sum_{i\in c} x_i \nonumber
\end{eqnarray}

\noindent where $c \in C$.  Since by definition $W'$ are centroids that minimize the
cluster assignments based on the previous $C$, $E(W',C) \le E(W,C)$ holds.

Having calculated the new centroids $W'$, new cluster
assignments $C'$ are made with respect to $W'$ that  minimize the
distance between each $x_i$ and its closest centroid $w'_{c(i)}$, yielding

\begin{eqnarray}
\sum_i  \frac{1}{2}  (x_i - w'_{c'(i)})^2 & \leq & \sum_i
\frac{1}{2} (x_i - w'_{c(i)})^2 \nonumber
\end{eqnarray}

\noindent which means $E(W',C') \leq E(W', C)$.  Thus, we've shown the
quantization error decreases at every iteration by $E(W',C') \leq
E(W',C) \leq E(W,C)$.  Finally, because there are a finite number of
data points and cluster assignments, k-means must converge.

\subsection{Convergence of Pk-means using Squared Distance}

We demonstrate that Pk-means converges using the squared distance
metric.  In Section \ref{sec:cross-corr-square-distance}, we establish
the relationship between cross correlation, the distance metric of
Pk-means, and squared distance.

Recall from Section \ref{sec:pkmeans} that Pk-means differs from
k-means in two ways.  First, Pk-means searches across all phase shifts $\{\tau\}_0^{d-1}$
when calculating distance between a time series $\vec{x}_i$ and centroid $\vec{w}$. Second, once Pk-means finds the centroid $\vec{w}_c(i)$ and phase shift $\tau_i$ that
minimizes distance, Pk-means horizontally adjusts $\vec{x_i}$ by
$\tau_i$ and produces $\vec{x_i}^{\tau_i}$.  We refer to the collective set of phase
adjustments with respect to centroids $W$ as $T$.  If phase
adjustments are done with respect to $W'$, we refer to these as
$T'$ and $\tau'_i$.   

In Figure \ref{alg:modified-kmeans}, distance calculation and cluster
assembly are listed as separate steps since algorithmically they
populate different data structures.  However, distance calculation
implies cluster composition since finding the closest centroid to
$\vec{x}_i$ effectively assigns that time series to a cluster. Thus,
phase shifts $T$ and cluster assignments $C$ are determined
simultaneously. We represent this relationship as tuple $(T,
C)$. Thus, Pk-means progresses from:

\begin{equation}
W \rightarrow (T, C) \rightarrow W' \rightarrow (T',C') \nonumber
\end{equation}

In order for Pk-means to converge, its quantization
error must also be non-increasing at each iteration. To incorporate
the effects of phase changes on the quantization error, we define the quantization error for Pk-means as

\begin{eqnarray}
 E(W,(T,C)) & = &\sum_{i=1}^n \frac{1}{2}(\vec{x}_i^{\tau_i} -
 \vec{w}_{c(i)})^2  \nonumber
\end{eqnarray}

\noindent Thus, our goal is to prove that $E(W', (T',C')) \leq E(W,
(T,C))$.  We do this by proving that $E(W',(T',C') \leq E(W', (T, C))
\leq E(W, (T,C))$.  Before we prove both expressions, we must derive
the value of $\vec{w}' \in W'$ under Pk-means.  This is the value that
minimizes $E(W',(T,C))$.  Taking the first derivative of this
expression with respect to $\vec{w}'$, setting equal to zero, and
solving for $\vec{w}'$, we get

\begin{eqnarray}
\label{Eq:vec_wprime} 
\vec{w}' & = & \frac{1}{|c|} \sum_{i \in c}
\vec{x}_i^{\tau_i}
\end{eqnarray}

\noindent The centroids $W'$ that minimize $E(W',(T,C))$ are obtained
by averaging the time series belonging to clusters $C$ but with the
added constraint that the data is phased according to $T$.  This is
exactly how Pk-means calculates the new centroids.  Since by
definition, $W'$ are the centroids that minimize the distance of the
data with respect to $(T,C)$, $E(W', (T, C)) \leq E(W, (T,C))$.

After $W'$ are calculated, we find the $T'$ and $C'$ that minimize the
distance of each $x_i$ to its closest centroid $\vec{w}'_{c'(i)}$.
From this, $E(W', (T',C')) \leq E(W', (T, C))$ follows.  Thus, we have
shown that the quantization error is non-increasing under the phase
changes of Pk-means.

\subsection{Cross Correlation versus Squared Distance}
\label{sec:cross-corr-square-distance}

Having shown that Pk-means converges using squared distance, we show
that Pk-means converges using cross correlation by establishing the
relationship between cross correlation and squared distance.  Recall
the definition of cross correlation in Equation
\ref{Equ:CrossCorrelation}.  The following equation demonstrates that,
if the vectors are normalized, squared distance and cross correlation
are inversely related.

\begin{eqnarray}
 \frac{1}{2}(\vec{x} - \vec{y})^2 & = & \frac{1}{2}(\vec{x}\cdot\vec{x} - 2\vec{x}\cdot\vec{y} + \vec{y}\cdot\vec{y}) \nonumber\\
                                  & = & \frac{1}{2}(1 - 2\vec{x}\cdot\vec{y} + 1) \nonumber\\
                                  & = & 1 - \vec{x}\cdot\vec{y} \nonumber\\
                                  & = & 1 - r^2_{xy} \nonumber
\end{eqnarray}

We show that using cross correlation does not impact the calculation of the centroids $W'$ by substituting the (negated) definition of cross correlation into expression $E(W',(T,C))$:

\begin{eqnarray}
 \sum_i \sum_{t=0}^{d-1} - w'_{c(i)}(t) \cdot x_i(t-\tau_i) & = & \nonumber \\
 \sum_i (\frac{1}{2}(\vec{x}_i^{\tau_i} - \vec{w}'_{c(i)})^2 - 1) & = & \nonumber \\
\sum_i  (\frac{1}{2}(\vec{x}_i^{\tau_i} - \vec{w}'_{c(i)})^2) - n & = & E(W',(T,C)) - n \nonumber 
\end{eqnarray}

\noindent Cross correlation changes the expression of $E(W',(T,C))$ only by the
subtraction of n.  Thus, centroid calculation remains unchanged under cross correlation, and the convergence of Pk-means
using cross-correlation follows from the proof of Pk-means convergence
using squared distance.


%



\section{Data}
\label{sec:data}

Before moving to our experimental results, we describe the four time
series data sets we experimented with.  Three are non-astrophysics
data sets.  The fourth data set is our motivating application:
light-curve data from the astrophysics domain.  We briefly describe
the three non-astrophysics data sets, and discuss in-depth the source,
generation and pre-processing of our light-curve data.

\subsection{Non-Astrophysics Time Series Data}

The first non-astrophysics time series data set, referred to as {\bf
nose} was generated by the Walt Laboratory at Tufts University.  Each
example was generated in the laboratory by passing a vapor over a succession of sensors.  The full data set contains eight classes (vapors) with fifteen examples each.  The length of each time series is eighty time points. 

The second data set, referred to as {\bf mallat}, is a synthetic data
set generated by Mallat \citep{mallat1998} for the study of wavelets in signal
processing, and is available at the UCR time series data archive
\citep{ucr-archive}.  The full data set contains eight classes
numbering 300 examples each.  Each example is 1024 time points.

The third data set, referred to as {\bf landcover} was donated by
Dr. Mark Friedl of Boston University.  This data set consists of time
series of satellite observations of the earth's surface. There are a
total of 18 classes, with each example containing 135 time points.
Note that the class distribution is not uniform.

In the state provided to us, these data sets are perfectly
synchronized.  The reason they were selected for experimentation is
that they serve as examples of the shapes of time series for which
universal phasing does not work (see Section \ref{sec:challenge}).  We
simulate the phasing problem by randomly rephasing (rotating) each
time series.  We then universally phase each randomly-rephased curve
to create the universally-phased version of those data sets.  Thus,
our experiments are run on both randomly-phased and universally-phased
data.  The only pre-processing performed on these data sets was
z-score normalization, which was done prior to the random- and
universal-phasing.

\subsection{Light-Curve Data}

\begin{figure}[tb]
    \centering
    \mbox {
    \subfigure[Before folding] {
           \label{fig:folding-before}
       \includegraphics[height=1.5in]{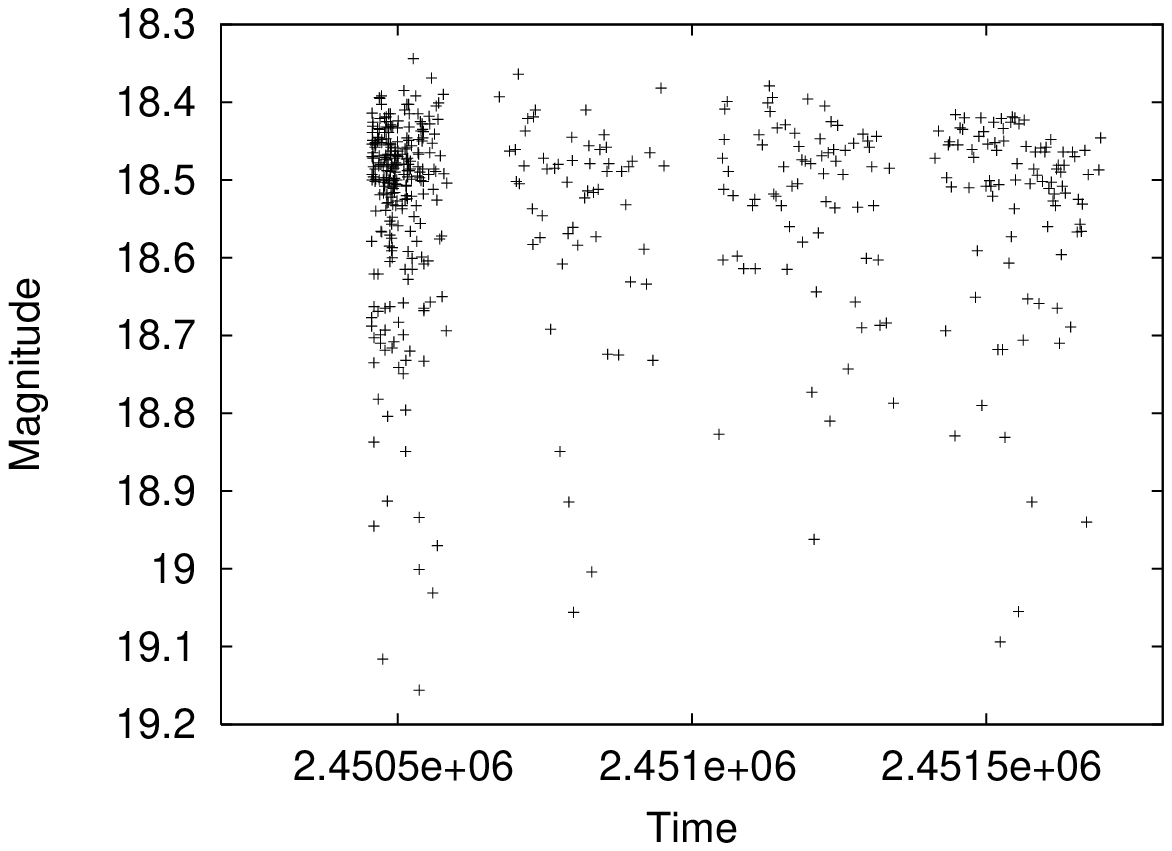} %
    } %
    \subfigure[After folding] {
           \label{folding-after}
       \includegraphics[height=1.5in]{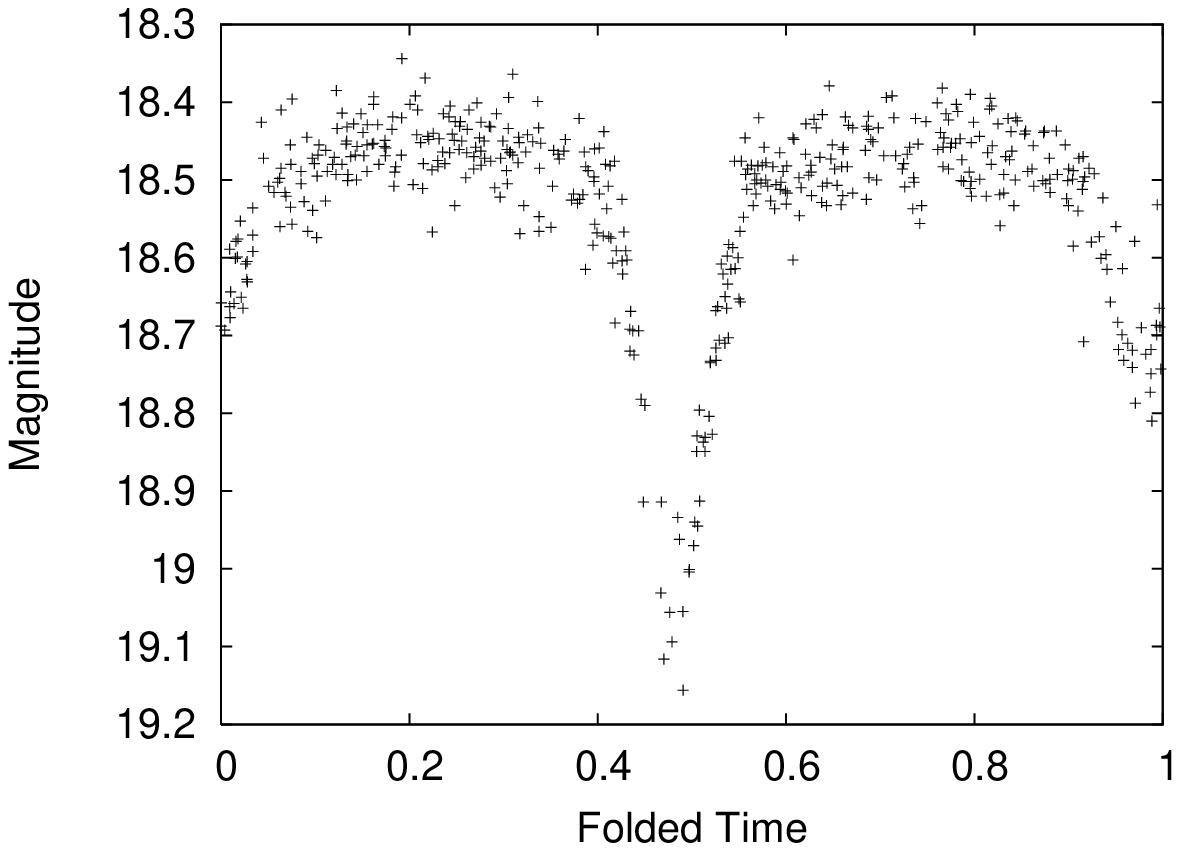} %
    } %
    }
    \caption{ An Eclipsing Binary light-curve before and after
    folding. Note that after folding the x-axis ranges from 0 to 1
    because only the non-integer part of the original time points are
    kept.}
    \label{fig:folding}
\end{figure}

The source of the light-curve data is the Optical Gravitational
Lensing Experiment Survey \citep{udalski97ogle}.  We refer to this
data as {\bf ogle}. The telescopes used for this survey capture CCD
images of the night sky over time.  Each digital image is carved into
``tiles'' and the stars in each tile are identified by number.  An
astronomer converts the series of image data for each star into a
light-curve - a real-valued series measuring the magnitude of light in
each image and its associated observational error. In order to keep
our algorithm generic to univariate time series data, PCAD uses only
the light magnitude measurement.  The result of not incorporating
observational errors into PCAD is that an otherwise typical
light-curve with noisy measurements may appear anomalous.  We take the
observational errors into account when the astronomers analyze the
ranked output from PCAD.  Because this increases the time an
astronomer must spend analyzing PCAD output, in future work we will
automate the handling of light-curves with high observational errors.

Because we work with periodic stars, we are able to transform the
photometric time series produced by the observing team of astronomers
into a single light-curve in which each period is mapped onto the
same time axis as follows:

\begin{equation}
  t' \equiv \Big\{ \frac{t-t_0}{T} \Big\}
\end{equation}

\noindent where $T$ is the period, $t_0$ is an arbitrary starting
point and the symbol $\{ \}$ represents the non-integer part of the
fraction.  This process produces a {\em folded light-curve} on an
x-axis of {\em folded time} that ranges from 0 to 1.  In the ogle data
sets, the lengths of the folded light-curves range between 200 to 500
time points. The purpose of folding is to produce a stronger, more
consistent signal.  Figure \ref{fig:folding} shows an example of an
Eclipsing Binary light-curve before and after folding.


The data are pre-processed according to a technique developed by
Protopapas et al. \citep{protopapas05b}.  We refer the reader to
\citep{protopapas05b} for an in-depth discussion of each technique and
its parameters. In brief, we conservatively pre-process the data using
spike removal, smoothing, interpolation and normalization.  Spike
removal and smoothing is used to eliminate time points that are
individually noisy, while preserving interesting features.  We use
interpolation because PCAD uses a FFT, which requires the time points
of each light-curve to be uniformly spaced.  Finally, we use z-score
normalization to ensure similarity calculations are meaningful.  The
pre-processing done on the light-curves is domain-specific, and not
part of the PCAD technique.  Whether time series from other domains
must undergo pre-processing is the decision of a domain expert for
that data set.

The ogle data are folded light-curves, which we simply
refer to as light-curves.  Our data sets consist of 1329 Cepheid
(CEPH), 2580 Eclipsing Binary (EB) and 5327 RR Lyrae (RRL)
light-curves, with each data set containing an unknown number of
anomalies. 

There are some important characteristics of these data sets that have
bearing on our experimental results.  The first is that the variance
on the EB data set is eight times the variance of either CEPH or RRL.
The second is that despite CEPH and RRL being distinct star
classes, the shapes of their light-curves are very similar (refer to
Figure \ref{fig:typical} for examples of both).  Indeed, domain
experts resort to other data sources to classify CEPH and RRL
light-curves correctly.  Due to the similarity in the shape of the
light-curves, time series anomaly detection methods, including ours,
have difficulty distinguishing between CEPH and RRL.

\section{Experiments}
\label{sec:Experiments}


Our main experimental goal is to justify the expense of PCAD by
comparing it to a variety of naive solutions.  The second goal is to
evaluate PCAD's performance as an approximation of a published anomaly
detection for light-curve data, and compare PCAD to other
approximations of this benchmark.

\subsection{Alternatives to PCAD}

Due to the lack of published alternatives to PCAD, we propose several
anomaly detection methods for handling large sets of unsynchronized
data that are both simpler and cheaper than PCAD. The first method
samples $k$ examples uniformly from the data set to generate a
reference set of ``centroids'', instead of running an expensive
algorithm like k-means or Pk-means.  One can then calculate the
outlier score using maximized cross correlation against this randomly
selected set.  We call this method RAND-CC.  Please note, that for
RAND-CC, sample size and centroid size are the same.

Another alternative is to set $k=1$ in Pk-means and generate a single
centroid from which the outlier scores are calculated.  This solution,
P1-MEAN, precludes finding local outliers, but may be an
acceptable solution for finding global outliers.  P1-MEAN has a cost
of $O(n)$ for both centroid and outlier score calculation.

Our final naive alternatives to PCAD are KMEANS-ED and KMEANS-CC.
K-means is a viable alternative to PCAD when run on universally-phased
data.  K-means is clearly cheaper than Pk-means because it lacks the
$\log{d}$ factor required by the FFT.  KMEANS-ED uses k-means to
generate centroids and uses Euclidean distance to calculate the
outlier scores.  KMEANS-CC also uses k-means to generate centroids,
and uses maximized cross correlation to calculate the outlier scores.

As a final point of comparison, we compare PCAD to the RI-DISCORD
method described in Section \ref{sec:related-work}. We implemented the
brute-force version of RI-DISCORD, as we are interested in comparing
performance rather than computational speed.  We also implement
RI-DISCORD using maximized cross correlation as the distance metric,
which is equivalent to the $O(n^2)$ measure of finding the minimum
distance over all possible rotations.  We define discords to be the
examples with the farthest nearest neighbor.

\subsection{Benchmark and Approximations}
\label{sec:benchmark}

On the ogle data set we compare PCAD's performance to a robust
solution called PN-MEANS \citep{protopapas05b}. This is a method
published in an astrophysics journal for light-curve data
specifically.  It is also an exhaustive version of PCAD where $k$ is
set to $n$ (the number of light-curves in the data set) and no
sampling is performed.  The anomaly score calculated for each
light-curve is the weighted average of its similarity to the other
$n-1$ light-curves.  The weights are Gaussian, with the sample mean
and standard deviation of the $n-1$ correlations plugged in as
parameters.  Note that PCAD differs from PN-MEANS in the weights used
for outlier score calculation.  While PN-MEANS uses Gaussian weights,
PCAD uses proportional cluster sizes as weights.  We decided against
using Gaussian weights for PCAD because PCAD calculates the outlier
scores from a small number of typical-looking centroids.  Gaussian
weighting is useful when the outlier score is calculated from a
larger, more diverse group of correlation scores.

By doing an exhaustive pair-wise comparison of the light-curves,
PN-MEANS is the most precise anomaly score that PCAD can
calculate. Hence, we consider PN-MEANS to be PCAD's
benchmark. However, because PN-MEANS sets $k=n$, it is $O(n^2)$.
Thus, PCAD is an improvement over PN-MEANS in terms of computational
expense, rather than quality of results.

We compare PCAD's ability to approximate PN-MEANS to two others
approximations of PN-MEANS.  The first method, called Protopapas\_n,
is an $O(n)$ approximation of PN-MEANS that calculates the outlier
score of each light-curve in relation to a single centroid, calculated
by averaging the examples in the data set \citep{protopapas05b}. This
is similar to P1-MEAN except that correlation rather than maximized
cross correlation is used as a distance metric.  The second method is
RAND-CC, which can also be thought of as an approximation of PN-MEANS.
While PN-MEANS builds a $n$ x $n$ similarity matrix and calculates the
outlier score by doing a weighted Gaussian averaging the columns,
RAND-CC builds a $r$ x $n$ similarity matrix where $r$ is the
cardinality of a randomly-selected subset of the data. The outlier
score calculation for RAND-CC is identical to PCAD in that it is a
weighted average of a time series' correlation to each of the
randomly-selected centroids, where the weights are the proportion of
the $n$ light-curves that are closest to each centroid. We also
created a second version of RAND-CC, called RAND-CC-GAUSS, which
differs from RAND-CC in that it uses Gaussian weighting to calculate
the outlier score instead of cluster proportions.  We compare PCAD to
Protopapas\_n, RAND-CC, and RAND-CC-GAUSS to show that PCAD is the
better approximation of PN-MEANS.

\subsection{Summary of Experimental Goals}

Having described the methods to which we compare PCAD, we summarize
our experimental goals as follows:

\begin{itemize}
  \item Evaluate PCAD's effectiveness in
comparison to the alternatives described above (KMEANS-ED,
KMEANS-CC, P1-MEAN, RAND-CC and RI-DISCORD) on randomly-phased data that
contains a known number of anomalies.
  \item Evaluate the same methods on the universally-phased versions of the data.
  \item Determine whether PCAD is a better approximation of PN-MEANS than either Protopapas\_n, 
RAND-CC, and RAND-CC-GAUSS on light-curve data containing an unknown number of anomalies.
 \item Understand the effect of parameter $k$ and sampling on PCAD's rankings.
 \item Provide an astrophysicist's analysis of the anomalies found by PCAD 

\end{itemize}

\section{Results}
\label{sec:results}

\subsection{Data with Known Anomalies}
\label{sec:known-anom}

We create data sets with known numbers of anomalies from each of the
four data sets we introduced in Section \ref{sec:data}. To create data
sets with global anomalies, we mix examples from three classes
together.  The first two classes are the ``normal'' classes.  They are
similar classes, and comprise 95\% of the synthetically-mixed data
set.  The remaining 5\% of the data are instances from an ``outlier''
class, which is a class that is dissimilar to the other two.  We
selected our class mixes either through domain expert advice or visual
inspection of the data. For the nose data set, 'hep' and 'tol' were
selected as the normal classes, and 'decoh' as the outlier class. For
the landcover data set, we selected 'Woody Savanna', and 'Cropland' as
the normal classes, and 'Snow-Ice' as the outlier class.  For the
mallat data set, classes 3 and 6 are the normal classes, and class 7
is the outlier class.  For the ogle data set, CEPH and RRL are the
normal classes, and EB is the outlier class.  The size of the
synthetically-mixed global anomaly data sets are 32, 210, 1050 and
1050 for nose, mallat, landcover and ogle respectively.

To create data sets with local outliers, we mix examples from four classes. Two classes are ``normal'' classes
and comprise 95\% of the data.  However, these two classes are
dissimilar and should not cluster together.  The remaining two ``outlier''
classes comprise 5\% of the data.  Each outlier class is similar to
one of the normal classes, and dissimilar to the other. Thus, each
outlier class should be a local outlier with respect to one of the
normal classes only.  For the nose data set, 'tol' is a local outlier
with respect to 'hep, and 'decoh' is a local outlier with respect to
'diesel'. For the mallat data set, classes 3 and 6 form one normal/outlier pair,
and classes 7 and 2 form another.  For the landcover data set, Croplands/Woody
Savanna form one pairing, and Open Shrubland/Snow-Ice another.
Because the ogle data set contained only three classes, it could not
be used to create a synthetically-mixed data set for local outliers.  The size of the
synthetically-mixed local anomaly data sets are 32, 210, and 1050 for
nose, mallat, and landcover respectively.

Because we know that the true number of anomalies in each data set is
$m$, we look at the top $m$ entries in each method's output and
measure precision.  Because there are $m$ true anomalies and $m$
reported anomalies in these experiments, precision and recall are
equal.  Thus, we report only one number.  We performed ten iterations
of each experiment.  In each iteration, we create a data set in the
proportions described above via random sampling. For the global
outlier experiments, we experimented with values of $k$ between 1 and
7, and selected the optimal $k$ values for k-means and Pk-means using
BIC.  For RAND-CC, we set the reference set or ``centroid'' size to
the value returned by BIC for Pk-means.  For example, if, for a given
experimental iteration, PCAD calculated the outlier score against four
centroids, then RAND-CC will calculate its outlier score against four
randomly-sampled light-curves.

For the local anomaly experiments, we set $k=2$ to force the evaluation of
local outliers with respect to two clusters.  P1-MEAN is omitted
from the local outlier experiments because it calculates only one
centroid.  RI-DISCORD is also omitted because it finds only global outliers.

We also demonstrate the effect of sample size on the results of
KMEANS-ED, KMEANS-CC, P1-MEAN, and PCAD by running both k-means and
Pk-means with sample sizes that are 5, 10, 20, 30, 40, 50 and 100\% of
the synthetically-mixed data sets.  Because the synthetically-mixed
nose data set is too small to be sampled (32 examples), we run
Pk-means on all thirty-two examples.  We ensure that the same centers
- in one case randomly-phased, in the other universally-phased -
initialize k-means and Pk-means for each experimental iteration.  We
performed the random-phasing by choosing an arbitrary time point in
each time series, and rotating the time series such that this chosen
time point becomes the starting point (time zero). The ogle data set
is naturally unsychronized, and has no need for random phasing.  Thus,
experiments labeled 'rand' for ogle refer to the time series in their
original form.  Universal phasing is performed according the method
specified in \citep{protopapas05b}.

\begin{figure}[tbp]
    \centering
     \mbox{
      \subfigure[landcover - rand] {
           \label{landcover-rand-gbl}
	   \includegraphics{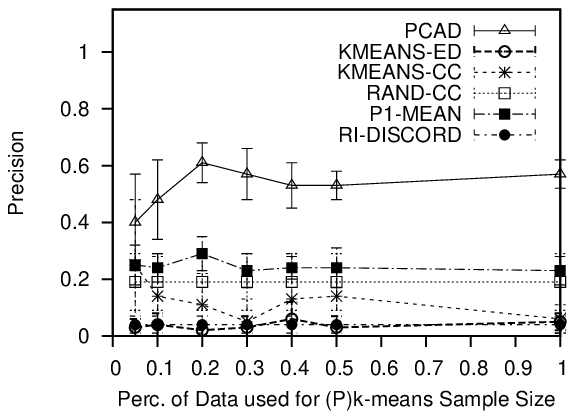} %
      } %
      \subfigure[landcover - univ] {
           \label{landcover-univ-gbl}
	   \includegraphics{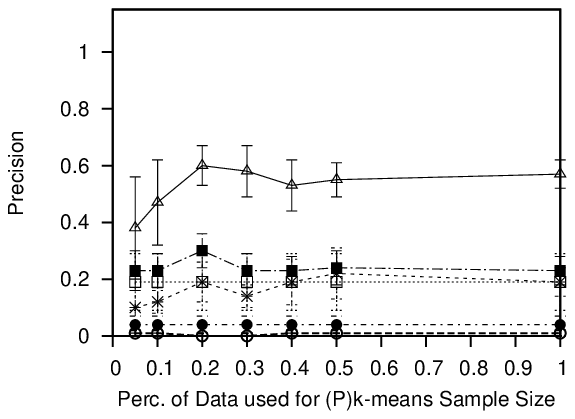} %
      } %
     }
     \mbox{
      \subfigure[mallat - rand] {
           \label{mallat-rand-gbl}
	   \includegraphics{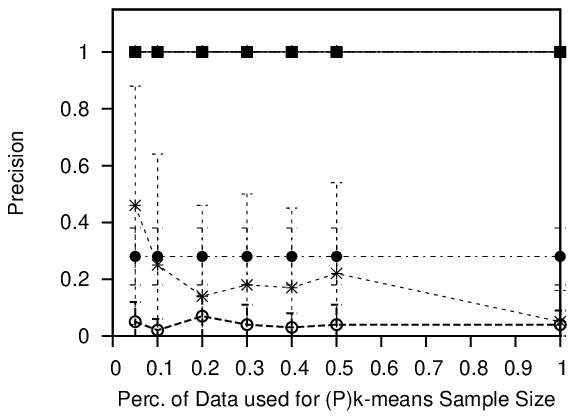} %
      } %
      \subfigure[mallat - univ] {
           \label{mallat-univ-gbl}
	   \includegraphics{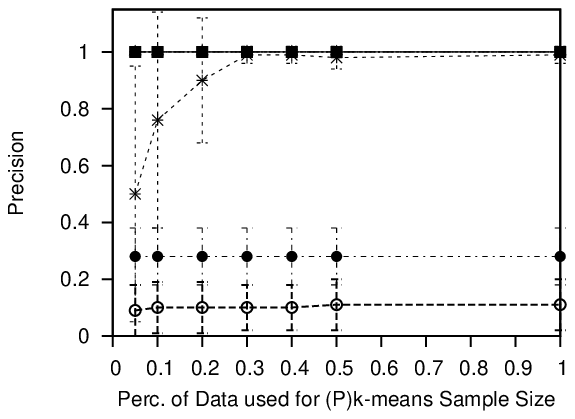} %
      } %
     }
     \mbox{
      \subfigure[ogle - rand] {
           \label{ogle_all-rand-gbl}
	   \includegraphics{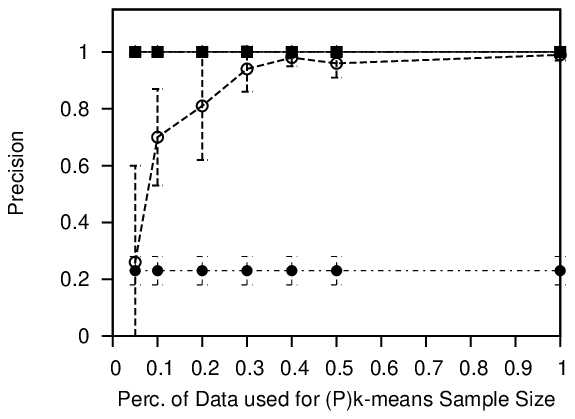} %
      } %
      \subfigure[ogle - univ] {
           \label{ogle_all-univ-gbl}
	   \includegraphics{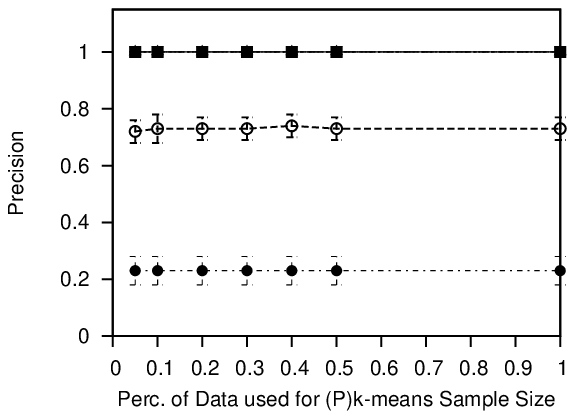} %
      } %
     }
    \caption{Precision vs. sample size for each data set synthetically-mixed to have {\bf global } outliers.  The x-axis measure the percentage of the data set used in the sampling given to either k-means or Pk-means.  The left-hand column shows results for randomly-phased data.  The right-hand side shows results for universally-phased data.  Note that because RAND-CC and RI-DISCORD do not have a (P)k-means subroutine, their results do not vary over the x-axis.  }
    \label{fig:syn-gbl}
\end{figure}

\begin{figure}[tbp]
    \centering
     \mbox{
      \subfigure[landcover - rand] {
           \label{landcover-rand-lcl-0}
	   \includegraphics{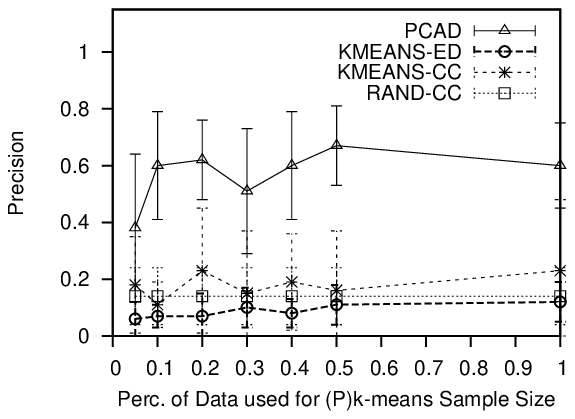} %
      } %
      \subfigure[landcover - univ] {
           \label{landcover-univ-lcl-0}
	   \includegraphics{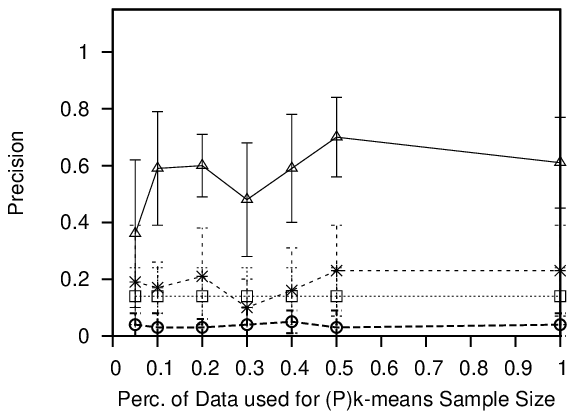} %
      } %
     }
     \mbox{
      \subfigure[mallat - rand] {
           \label{mallat-rand-lcl-0}
	   \includegraphics{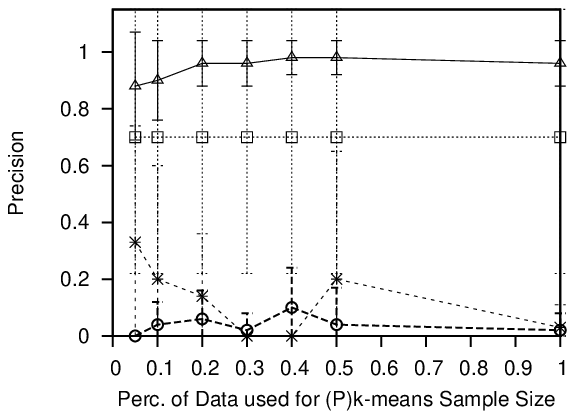} %
      } %
      \subfigure[mallat - univ] {
           \label{mallat-univ-lcl-0}
	   \includegraphics{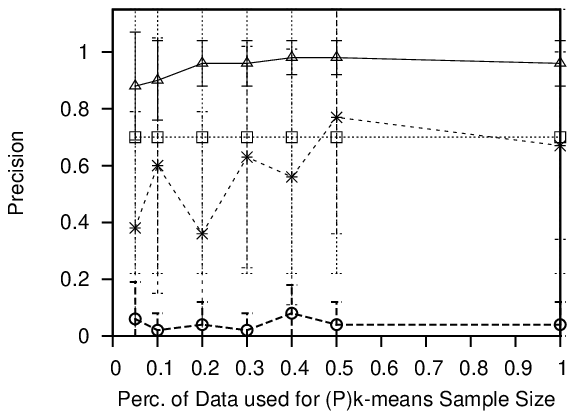} %
      } %
     }%
   \caption{Precision vs. sample size for each data set synthetically-mixed to have {\bf local } outliers.  Results are for local outliers with respect to {\bf centroid 1}. The x-axis measure the percentage of the data set used in the sampling given to either k-means or Pk-means.  The left-hand column shows results for randomly-phased data.  The right-hand side shows results for universally-phased data.  Note that because RAND-CC does not have a (P)k-means subroutine, its results do not vary over the x-axis.}
    \label{fig:syn-lcl-0}
\end{figure}

\begin{figure}[tbp]
    \centering
     \mbox{
      \subfigure[landcover - rand] {
           \label{landcover-rand-lcl-1}
	   \includegraphics{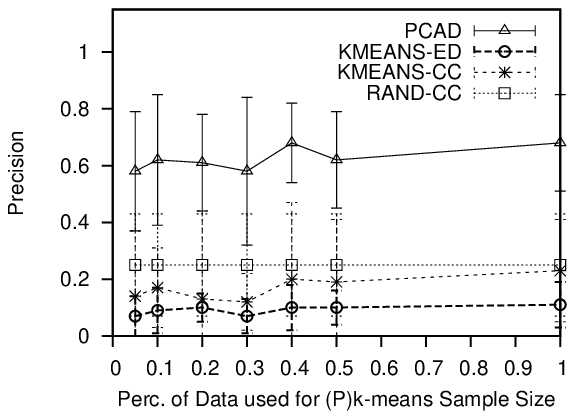} %
      } %
      \subfigure[landcover - univ] {
           \label{landcover-univ-lcl-1}
	   \includegraphics{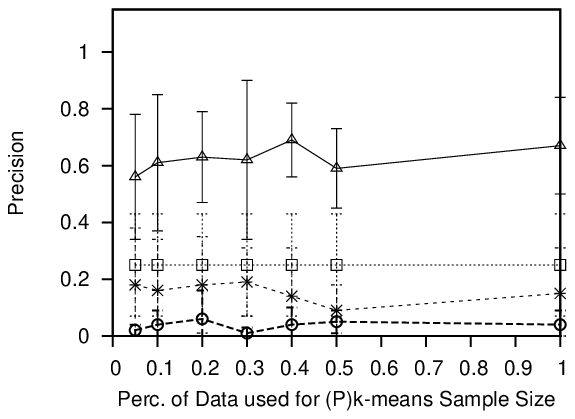} %
      } %
     }
     \mbox{
      \subfigure[mallat - rand] {
           \label{mallat-rand-lcl-1}
	   \includegraphics{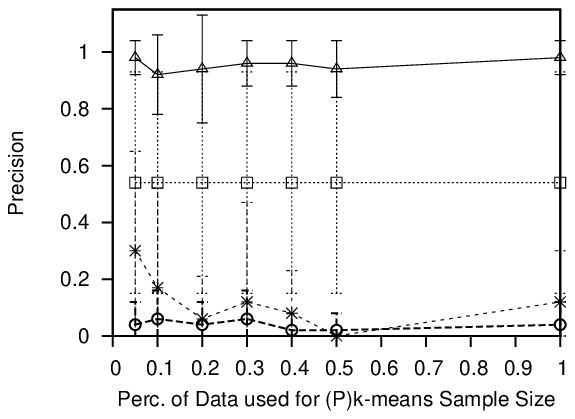} %
      } %
      \subfigure[mallat - univ] {
           \label{mallat-univ-lcl-1}
	   \includegraphics{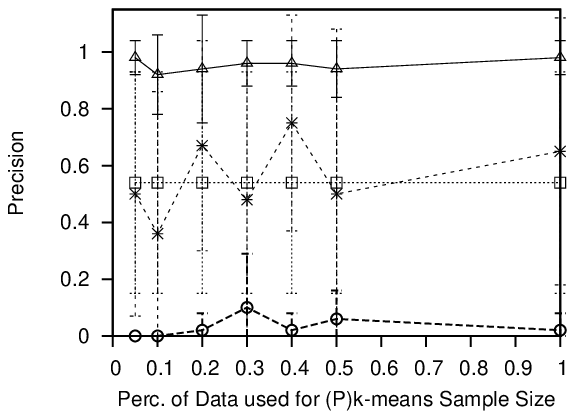} %
      } %
     }%
   \caption{Precision vs. sample size for each data set synthetically-mixed to have {\bf local } outliers.  Results are for local outliers with respect to {\bf centroid 2}. The x-axis measure the percentage of the data set used in the sampling given to either k-means or Pk-means.  The left-hand column shows results for randomly-phased data.  The right-hand side shows results for universally-phased data.  Note that because RAND-CC does not have a (P)k-means subroutine, its results do not vary over the x-axis.}
    \label{fig:syn-lcl-1}
\end{figure}

\begin{table}[th]
\centering
\resizebox{0.95\columnwidth}{!}{
\begin{tabular}{|r|r|c|c|c|c|c|c|}
\hline
DATA & GBL/LCL & KMEANS-ED & KMEANS-CC & RAND-CC & P1-MEAN & PCAD & RI-DISCORD \\
\noalign{\smallskip}\hline\noalign{\smallskip}
 gbl & rand  & 0.00 $\pm$ 0.00 & 0.45 $\pm$ 0.44 & 0.75 $\pm$ 0.26 & 1.00 $\pm$ 0.00 & 1.00 $\pm$ 0.00 & 0.45 $\pm$ 0.16\\
 gbl & univ  & 0.00 $\pm$ 0.00 & 0.50 $\pm$ 0.00 & 0.75 $\pm$ 0.26 & 0.95 $\pm$ 0.16 & 1.00 $\pm$ 0.00 & 0.45 $\pm$ 0.16\\
 lcl.1 & rand  & 0.10 $\pm$ 0.32 & 0.33 $\pm$ 0.50 & 0.20 $\pm$ 0.42 &  & 0.20 $\pm$ 0.42 & \\
 lcl.1 & univ  & 0.00 $\pm$ 0.00 & 0.00 $\pm$ 0.00 & 0.20 $\pm$ 0.42 &  & 0.40 $\pm$ 0.52 & \\
 lcl.2 & rand  & 0.00 $\pm$ 0.00 & 0.60 $\pm$ 0.52 & 0.20 $\pm$ 0.42 &  & 0.70 $\pm$ 0.48 & \\
 lcl.2 & univ  & 0.10 $\pm$ 0.32 & 0.10 $\pm$ 0.32 & 0.20 $\pm$ 0.42 &  & 0.60 $\pm$ 0.52 & \\
\noalign{\smallskip}\hline
\end{tabular}
}
\caption{All precision results for the nose data set.  'gbl','lcl.1', and 'lcl.2' refer to results on global outliers, and local outliers with respect to clusters 1 and 2. }
\label{tbl:syn-nose}
\end{table}

Figures \ref{fig:syn-gbl}, \ref{fig:syn-lcl-0}, and
\ref{fig:syn-lcl-1} plot the mean precision versus sample size on the
randomly- and universally-phased data sets of landcover, mallat and
ogle.  All results for nose are reported in Table \ref{tbl:syn-nose}.
Because, RAND-CC and RI-DISCORD do not use k-means or Pk-means to
generate centroids, we show their
results as a straight line in each plot.  Also, because RAND-CC and
RI-DISCORD use maximized cross correlation to calculate the outlier
score, the results for RAND-CC and RI-DISCORD are identical for
randomly- and universally-phased data.  PCAD and P1-MEAN also use
maximized cross correlation, but are not guaranteed to have identical results between
randomly- and universally-phased data because the intermediate step of
centroid calculation is done with differently-phased time
series. Nevertheless, the results are very similar.  KMEANS-CC's and KMEANS-ED's results change more dramatically
between randomly- and universally-phased data.

Our first observation is that PCAD performs best overall.  On global anomaly experiments, it performs perfectly on the nose, mallat and ogle for both randomly- and universally-phased data.  On all other experiments, local and global, it is the best performing method with one exception (the 'lcl.1' experiment on the randomly-phased nose data).

The methods that perform most poorly are KMEANS-ED and RI-DISCORD.
The poor performance of KMEANS-ED is expected considering there is no
attempt to phase the data in either the centroid or outlier score
calculation.  Thus, its results are poor for both randomly and
universally-phased data. Surprisingly, KMEANS-ED shows good results on
the synthetically-mixed ogle data set for global outliers.  There are
two reasons for this.  The first is that it is very easy to
distinguish EB, the minority light-curve in this data set, from both
CEPH and RRL.  This demonstrates that even a poorly-formed centroid
may be good enough for the task of anomaly detection.  The second
reason why KMEANS-ED has good results on ogle is that the ogle data
set universally-phases well.  In Figure \ref{fig:typical}, notice that
a typical light-curve from each star class has a global maximum.  We
hypothesize that KMEANS-ED has a better performance on randomly-phased
data at higher sample sizes because those centroids tend towards a
horizontal line.  For global anomaly detection on this
synthetically-mixed ogle data set, a horizontal line is a sufficient
model for distinguishing EB from CEPH and RRL!

RI-DISCORD's performance is not a reflection upon its ability to
handle unphased data.  Its poor overall performance is due to the
discord definition.  These results demonstrate that the farthest nearest
neighbor definition is not sufficient for finding the minority classes
in these data sets.  The results may improve upon
changing the discord definition to the farthest distance to its $k$-th
nearest neighbor, but one must know how to set $k$ in advance.

The global anomaly results for KMEANS-CC on ogle also demonstrate that
k-means's centroids may be good enough for anomaly detection.  Those
results are perfect for the same reasons described for KMEANS-ED.
However, for the data sets that do not universally-phase well, its
precision results on randomly- and universally-phased data are
similarly weak. This is also true for local anomaly results on mallat,
nose and landcover.  The results for KMEANS-CC and KMEANS-ED show that
bad centroids may be good enough for some data sets, but the centroid
generation of k-means is not robust in general.  It is certainly not
sufficiently robust for local anomaly detection.

RAND-CC has perfect precision on global anomaly detection on the ogle
and mallat (its line is superimposed on top of PCAD's and P1-MEANS's
in the figures), but performs below PCAD on the landcover and
nose data sets for global anomaly detection.  It always performs below
PCAD for local anomaly detection.  Also, RAND-CC always has a higher
standard deviation on its results compared to PCAD.  RAND-CC's random
selection of centroids averts the bad centroid problem of KMEANS-CC
and KMEANS-ED, but leads to a higher variation
in the quality of anomalies found.  Random centroids are
clearly inferior to Pk-means-generated centroids for local anomaly detection.

P1-MEAN is closest in performance to PCAD.  Its global anomaly results
are either perfect or nearly perfect on nose, mallat and ogle.
However, PCAD has better performance on landcover.  The landcover data
set is the hardest of the four synthetically-mixed sets.  The data
have a high signal-to-noise ratio and there is a lot of variance
within each class.  Under these circumstances, PCAD's multiple
centroids model the data better than P1-MEAN's single centroid.

\subsection{Light-curve Data with Unknown Anomalies}

In the next sections we move on to experiments with data sets CEPH, EB, and 
RRL.  Because these data sets contain an unknown number of 
anomalies, it is not possible to report precision.  The reporting of
precision would require our domain experts to examine the top $m$ list
of outliers from each experimental iteration, an act which is too
time-consuming.  Our experts examine a single set of results in Section
\ref{astro-review}.

We measure how well PCAD and Protopapas\_n (described in Section
\ref{sec:benchmark}) approximate PN-MEANS by measuring the change in
rank of PN-MEANS's top $m$ anomalies.  Specifically, we compare the
rankings of PN-MEANS's $m$ most anomalous light-curves to the rankings
produced by either PCAD or Protopapas\_n, seeking a minimal change in
rank between the two methods.  For example, if $m=3$ and PN-MEANS and
PCAD rank the top three light-curves at 1, 2, and 3, and 1, 6, and 7,
respectively, then these light-curves have a rank change of $|1-1|=0$,
$|2-6|=4$, and $|3-7|=4$ respectively.

\begin{figure}[tbp]
    \centering
	\mbox{
  	  \subfigure[CEPH - PCAD] {
      	     \label{fig:rc-ceph}
	     \includegraphics[width=2in]{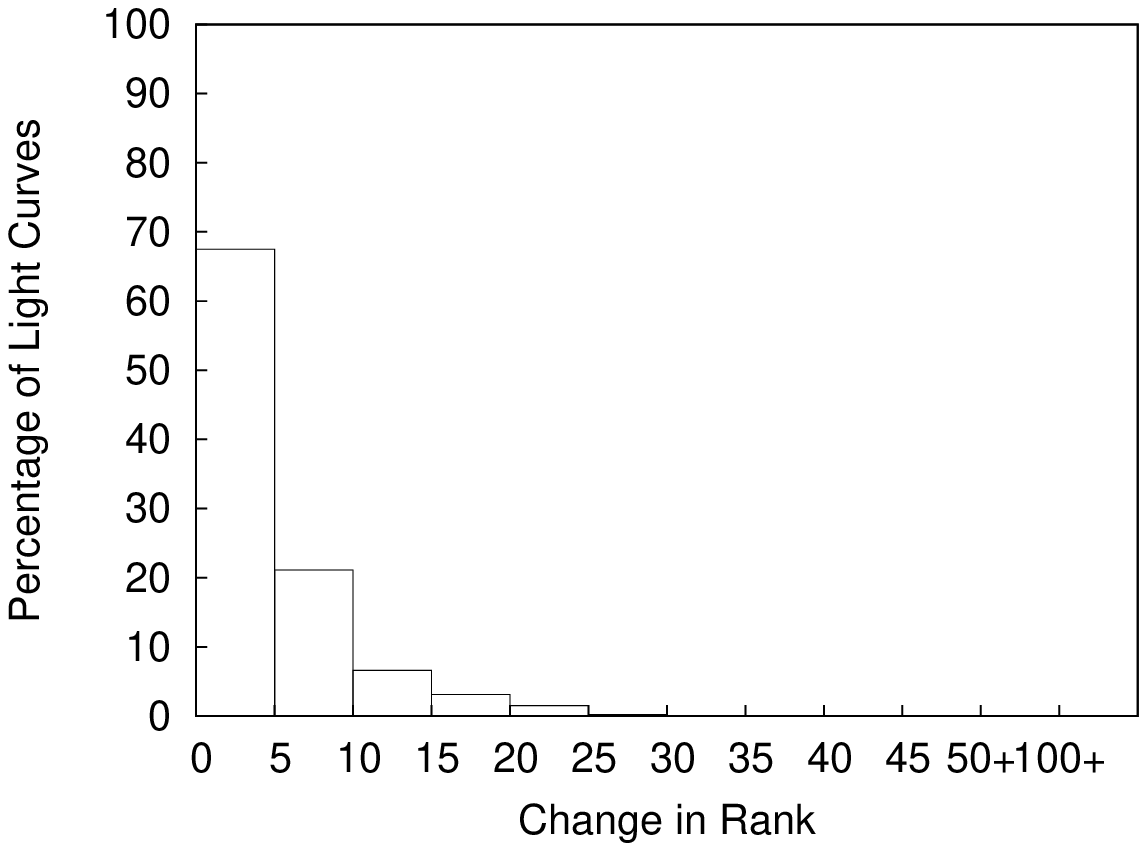} %
           } %
  	  \subfigure[CEPH - Protopapas\_n] {
      	     \label{fig:rc-ceph-pp1}
	     \includegraphics[width=2in]{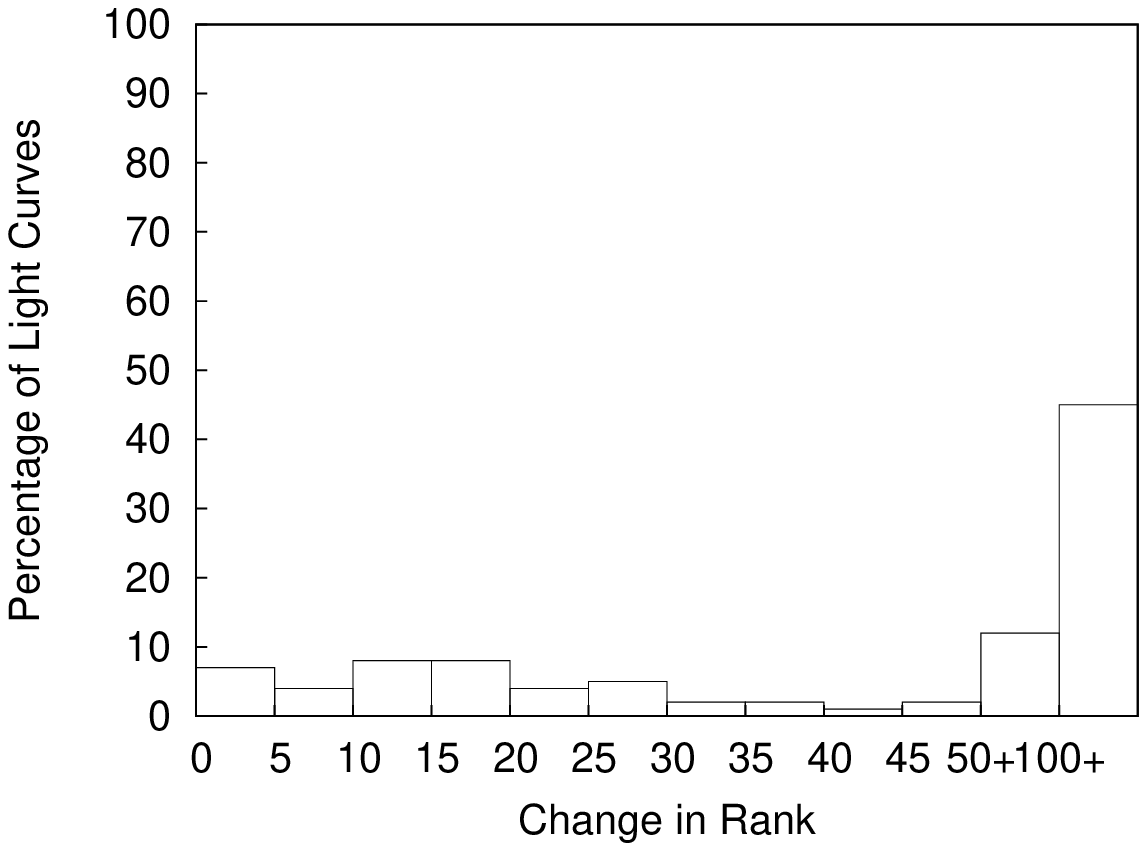} %
           } %
	}
	\mbox{
  	  \subfigure[EB - PCAD] {
      	     \label{fig:rc-eb}
	     \includegraphics[width=2in]{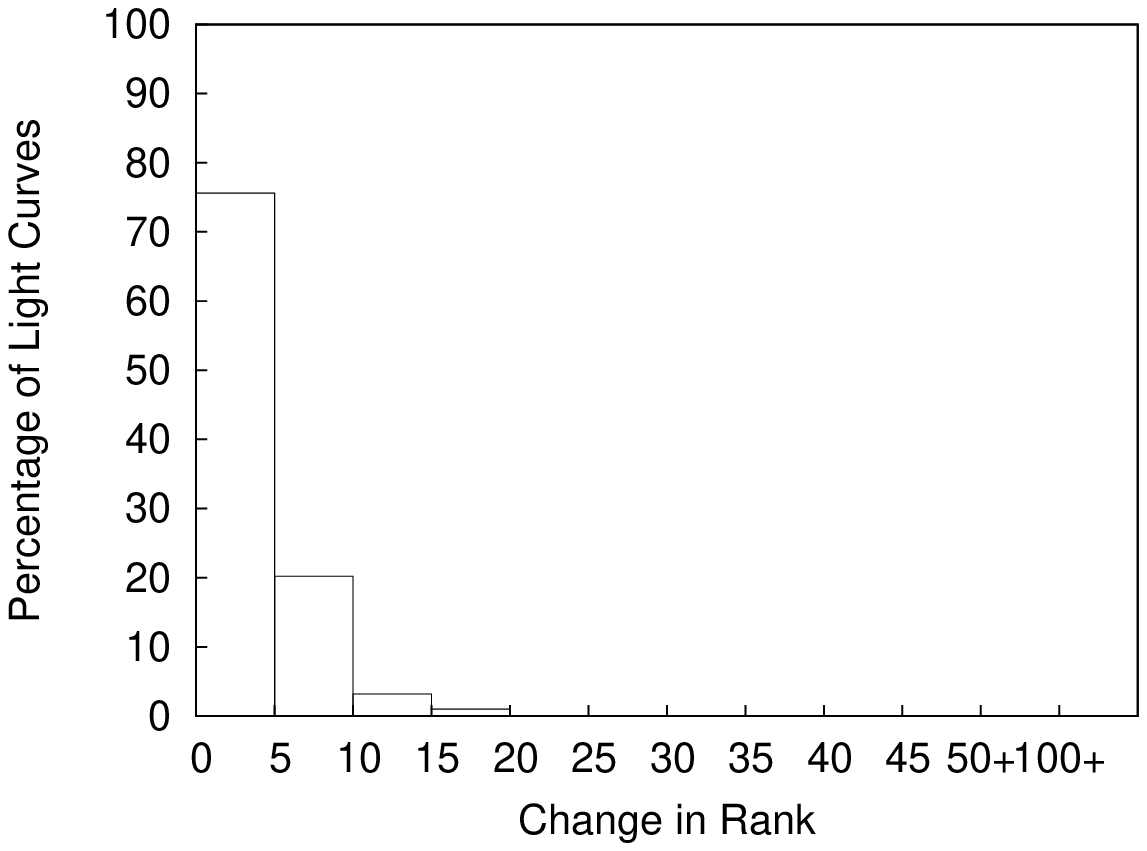} %
           } %
  	  \subfigure[EB - Protopapas\_n] {
      	     \label{fig:rc-eb-pp1}
	     \includegraphics[width=2in]{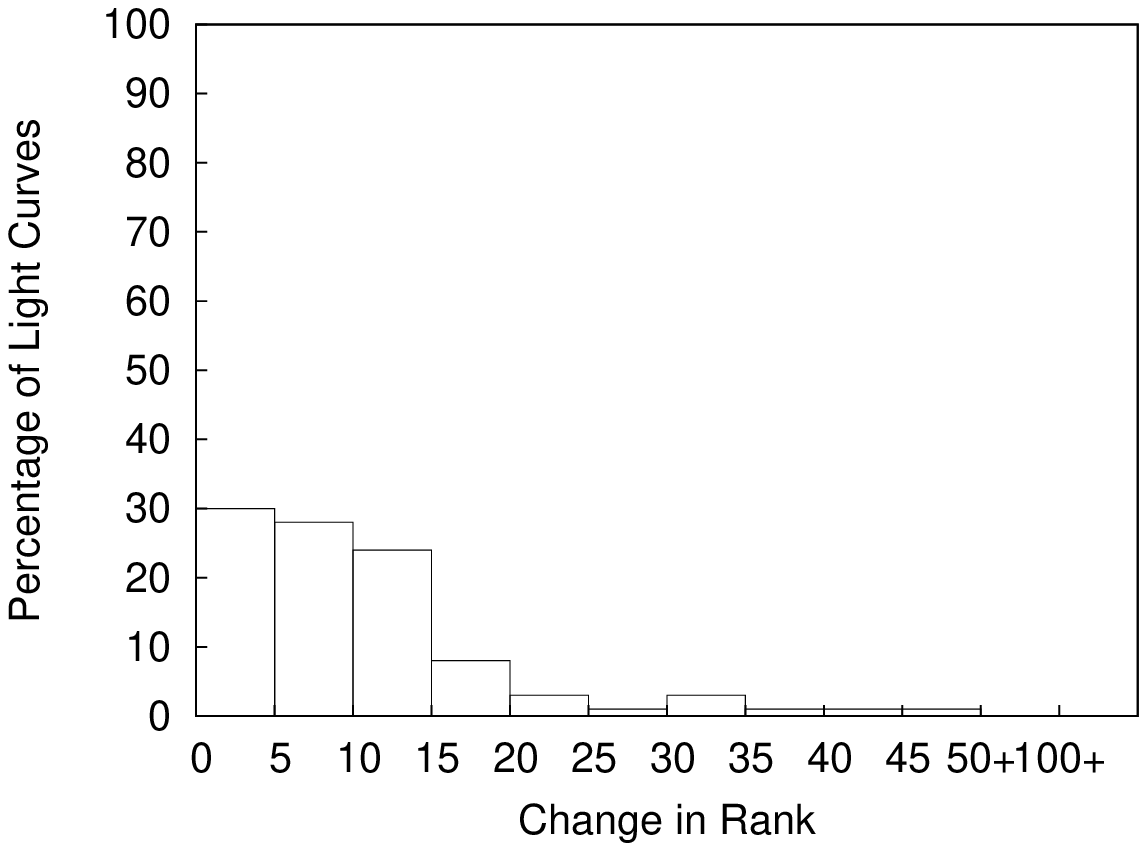} %
           } %
	}
	\mbox{
  	  \subfigure[RRL - PCAD] {
      	     \label{fig:rc-rrl}
	     \includegraphics[width=2in]{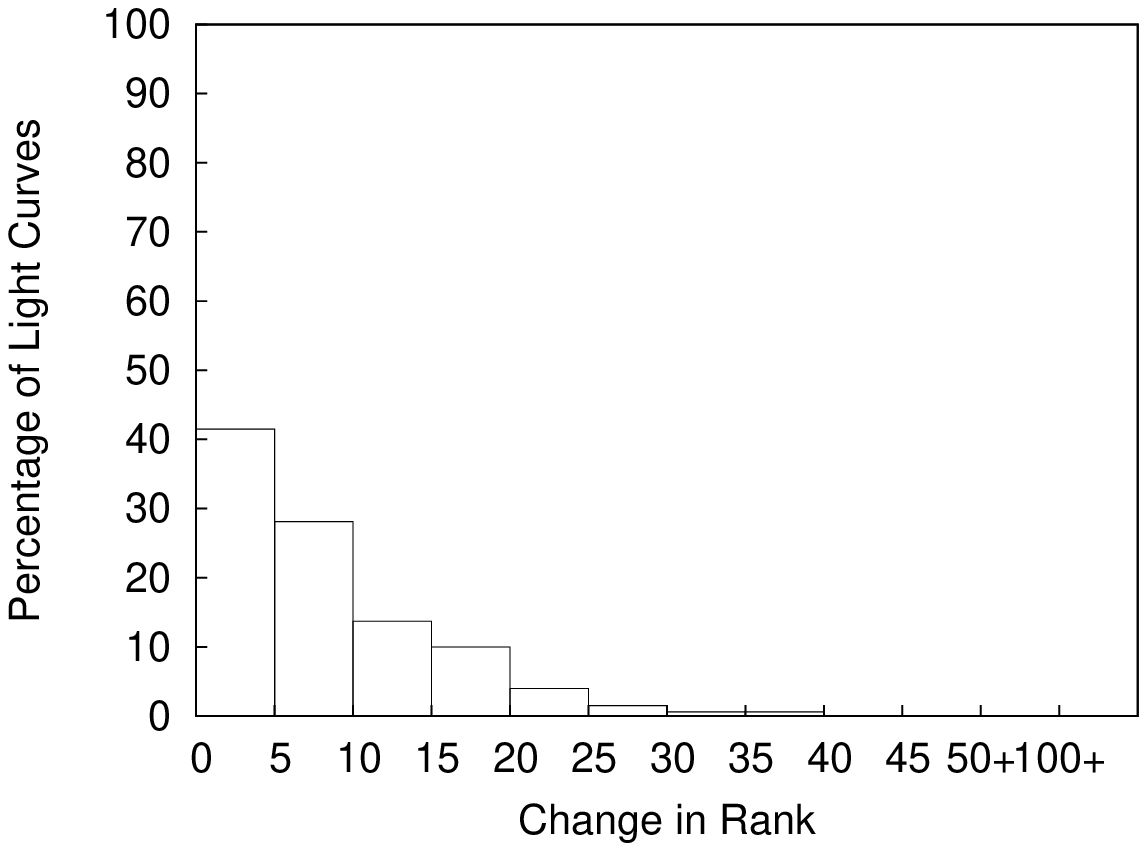} %
           } %
  	  \subfigure[RRL - Protopapas\_n] {
      	     \label{fig:rc-rrl-pp1}
	     \includegraphics[width=2in]{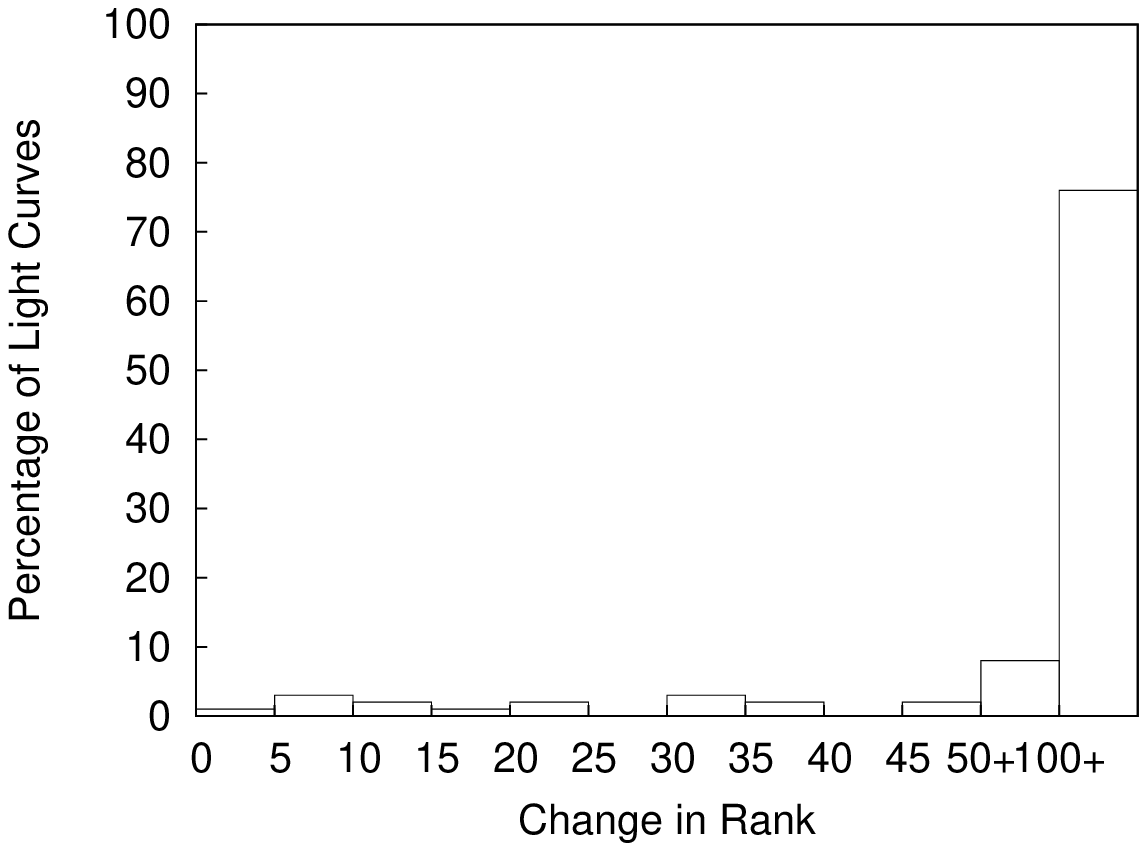} %
           } %
	}
        \caption{ Histograms of rank change comparing the rankings of both PCAD and Protopapas\_n to the benchmark PN-MEANS.}
    \label{fig:global-rc}
\end{figure}

We first show results for the detection of global anomalies on CEPH,
EB, and RRL.  We experimented with values of $k$ in $[1,10]$ and
sample sizes $[100,1000]$ at ten iterations each.  We then set PCAD to
return a ranked list of 100 light-curves, and average the rankings of
the ten iterations with the highest BIC.  Figure \ref{fig:global-rc}
shows our histograms of rank change that compare PCAD and
Protopapas\_n to benchmark PN-MEANS on sample size 500 (chosen
arbitrarily).  In all cases, PCAD's anomaly rankings are much closer
to those of PN-MEANS than Protopapas\_n's rankings.  Thus, PCAD is the
better approximation of PN-MEANS.  Protopapas\_n fares better on the
EB data set because it is a higher variance data set whose outliers are
more discernible.


\subsection{Effect of Sample Size}
\label{sec:ss-effect}

\begin{figure}[tbp]
    \centering
      \subtable[CEPH] {
\begin{tabular}[c]{|r|c|c|c|}
\hline
SS & PCAD & RND-CC & RND-CC-G \\
\hline
20  &  & 6.29 $\pm$ 1.67 & 69.50 $\pm$ 76.25  \\
40  &  & 4.49 $\pm$ 1.12 & 20.86 $\pm$ 16.80  \\
60  &  & 4.26 $\pm$ 0.57 & 14.26 $\pm$ 10.70  \\
80  &  & 3.86 $\pm$ 0.51 & 7.82 $\pm$ 2.06  \\
100  & 4.19 $\pm$ 1.86 & 4.54 $\pm$ 0.60 & 13.68 $\pm$ 6.37  \\
200  & 2.22 $\pm$ 1.43 & 4.22 $\pm$ 0.43 & 8.37 $\pm$ 2.17  \\
400  & 1.97 $\pm$ 1.11 & 4.05 $\pm$ 0.19 & 8.05 $\pm$ 0.95  \\
600  & 1.20 $\pm$ 0.45 & 4.22 $\pm$ 0.16 & 8.82 $\pm$ 0.42  \\
800  & 1.04 $\pm$ 0.25 & 4.22 $\pm$ 0.18 & 10.77 $\pm$ 0.72  \\
1000  & 0.78 $\pm$ 0.38 & 4.27 $\pm$ 0.10 & 14.72 $\pm$ 0.51  \\
\noalign{\smallskip}\hline\noalign{\smallskip}
\end{tabular}      } %

      \subtable[EB] {
	\label{tbl:ss-eb}
\begin{tabular}[c]{|r|c|c|c|}
\hline
SS & PCAD & RND-CC & RND-CC-G \\
\hline
20  &  & 3.72 $\pm$ 0.80 & 8.96 $\pm$ 3.83  \\
40  &  & 2.90 $\pm$ 0.53 & 8.33 $\pm$ 1.74  \\
60  &  & 2.82 $\pm$ 0.41 & 7.63 $\pm$ 1.88  \\
80  &  & 2.44 $\pm$ 0.27 & 6.42 $\pm$ 1.15  \\
100  & 5.37 $\pm$ 1.01 & 2.73 $\pm$ 0.55 & 7.57 $\pm$ 2.31  \\
200  & 4.46 $\pm$ 0.73 & 2.50 $\pm$ 0.22 & 8.04 $\pm$ 1.49  \\
400  & 3.36 $\pm$ 1.27 & 2.28 $\pm$ 0.21 & 7.56 $\pm$ 1.22  \\
600  & 2.35 $\pm$ 0.86 & 2.25 $\pm$ 0.17 & 7.64 $\pm$ 0.24  \\
800  & 1.31 $\pm$ 0.35 & 2.31 $\pm$ 0.12 & 7.96 $\pm$ 0.70  \\
1000  & 1.29 $\pm$ 0.58 & 2.30 $\pm$ 0.09 & 8.34 $\pm$ 0.62  \\
\noalign{\smallskip}\hline\noalign{\smallskip}
\end{tabular}
      } %

      \subtable[RRL] {
	\label{tbl:ss-rrl}
\begin{tabular}[c]{|r|c|c|c|}
\hline
SS & PCAD & RND-CC & RND-CC-G \\
\hline
20  &  & 7.61 $\pm$ 1.17 & 10.62 $\pm$ 4.67  \\
40  &  & 7.13 $\pm$ 0.85 & 7.05 $\pm$ 2.21  \\
60  &  & 6.84 $\pm$ 0.66 & 6.26 $\pm$ 2.74  \\
80  &  & 6.88 $\pm$ 0.53 & 8.03 $\pm$ 2.60  \\
100  & 3.56 $\pm$ 0.75 & 6.87 $\pm$ 0.21 & 6.61 $\pm$ 1.28  \\
200  & 2.80 $\pm$ 0.88 & 6.73 $\pm$ 0.36 & 5.30 $\pm$ 0.67  \\
400  & 2.07 $\pm$ 0.33 & 6.72 $\pm$ 0.21 & 5.68 $\pm$ 0.57  \\
600  & 1.55 $\pm$ 0.27 & 6.82 $\pm$ 0.19 & 5.65 $\pm$ 0.54  \\
800  & 1.33 $\pm$ 0.40 & 6.78 $\pm$ 0.14 & 5.93 $\pm$ 0.63  \\
1000  & 0.99 $\pm$ 0.28 & 6.81 $\pm$ 0.13 & 5.88 $\pm$ 0.59  \\
\noalign{\smallskip}\hline\noalign{\smallskip}
\end{tabular}
      } %
    \caption{Comparison of mean rank change of the top 100
      light-curves of PCAD, RAND-CC and RAND-CC-GAUSS on CEPH, EB and
      RRL using varying sample sizes (SS).  For PCAD, sample size
      refers to the size of the data set Pk-means is run on.  Pk-means
      produces no more than ten centroids for outlier score
      calculation.  For RAND-CC and RAND-CC-GAUSS, sample sizes refers
      to the number of centroids that the outlier scores was
      calculated with.}
    \label{tbl:ss}
\end{figure}

We also compare the quality of approximation among PCAD, RAND-CC, and
RAND-CC-GAUSS (recall from Section \ref{sec:benchmark} that RAND-CC
and RAND-CC-GAUSS can be considered sampled versions of PN-MEANS).  In
Table \ref{tbl:ss}, we show the effects of sample size on both PCAD,
RAND-CC, and RAND-CC-GAUSS.  For all methods, it is important to
understand how the term sample size is used.  For PCAD, a sampling
of the data is used by Pk-means to generate a relatively small set of centroids.
For both RAND-CC methods, the samplings are used directly as the
centroids for comparison.  Thus, for the RAND-CC methods, sample size
and centroid size can be used interchangeably. For PCAD, sample size
and centroid size are distinct entities.  The centroid size of PCAD is chosen
by BIC.  In our experiments, the number of centroids that maximizes
BIC is never more than ten.

We compare PCAD to both RAND-CC methods on a wide range of sample
sizes (between 10 and 1000).  The purpose of testing on smaller sample
sizes is to compare PCAD and the RAND-CC methods when they use a
comparable number of centroids. We compare performance at larger
sample sizes to test the effect of directly converting the sampling
intended for Pk-means into the reference set of centroids used by
RAND-CC.  Of course, at larger sample sizes, the computational expense
of both RAND-CC methods increases, approaching $O(n^2)$ as the sample
sizes approaches the size of the data set.  Thus, it is not desirable
for RAND-CC to use too large a sample.

Table \ref{tbl:ss} shows average change in rank between each method
and PN-MEANS for the top 100 anomalies.  For PCAD, the results show
that increasing the sample size used by Pk-means improves PCAD's
ability to approximate PN-MEANS.  At a sample size of 1000, the
average rank change of the top 100 anomalies is less than 1.30 for all
star classes.  Both RAND-CC methods do not approach this level of
approximation, even at sample sizes (really, centroid sizes) of 1000.
RAND-CC-GAUSS, which differs from RAND-CC in using a Gaussian-weighted
outlier scoring function, shows extremely high variabilty at small
sample sizes.  This is because Gaussian weighting does not work well
with small sample sizes.  Another observation is that the average rank
change values do not decrease monotonically for RAND-CC-GAUSS, though
the trend is generally downward.  But despite this downward trend,
RAND-CC-GAUSS never outperforms PCAD, even at sample size 1000.  This
is particularly surprising for the CEPH data set, where 1000 examples
is approximately 75\% of the data.  Independently, we verified that
when the sample size approaches the actual size of the data, the rank
change does indeed converge to 0 for RAND-CC-GAUSS.  So, even with
75\% of the data set represented in the sample, the variability of the
sample causes RAND-CC-GAUSS to underperform PCAD as an approximation
method.  Our hypothesis for why both RAND-CC methods underperform is
that their outlier score calculations are exposed to the effects of
noise in the light-curves and variability in the sample.  Meanwhile,
PCAD is effective despite using small numbers of centroids because
Pk-means is able to effectively mitigate both effects and produce a
reliable set of centroids.

\subsection{Effect of Parameter $k$}
\label{sec:k-effect}

\begin{figure}[tb]
    \centering
    \includegraphics[angle=270]{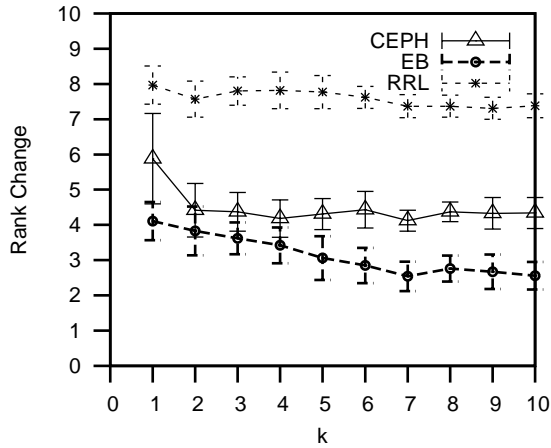} %
    \caption{Effect $k$ on PCAD.  The y-axis measure average change in rank from PN-MEAN over PCAD's top 100 anomalies.}
    \label{fig:k}
\end{figure}

We currently use BIC as a guideline for the selection of $k$.  Figure
\ref{fig:k} shows average change in rank between each PCAD and
PN-MEANS for the top 100 global anomalies over increasing values of
$k$. The results show that a wide range of $k$ does not greatly impact
the rankings. One might conclude from these results that the selection
of $k$ may be trivial, and one can choose $k=1$ and reduce the
computational cost of Pk-means.  However, our earlier experimental
results comparing PCAD to P1-MEAN (Figure \ref{fig:syn-gbl}) justified
our multi-centroid approach.  Furthermore, a choice of $k=1$
eliminates the possibility of finding local anomalies.

\subsection{Astrophysicists' Review of Anomalies}
\label{astro-review}

\begin{figure}[tbp]
    \centering
    \mbox {
      \subfigure[17.70123 (CEPH)] {
        \label{fig:17.70123}
	\includegraphics{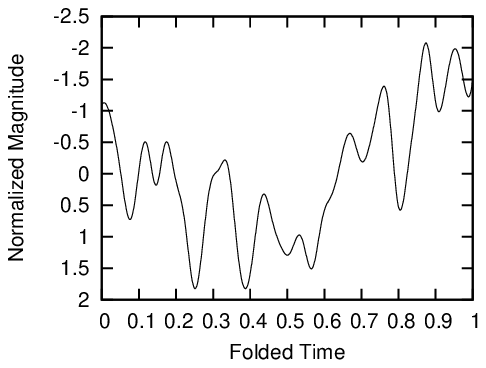} %
      } %
      \subfigure[13.184117 (CEPH)] {
        \label{fig:13.184117}
	\includegraphics{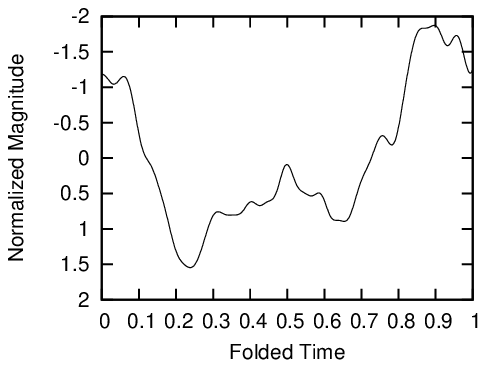} %
      } %
    }
    \mbox {
      \subfigure[20.54206 (CEPH)] {
        \label{fig:20.54206}
	\includegraphics{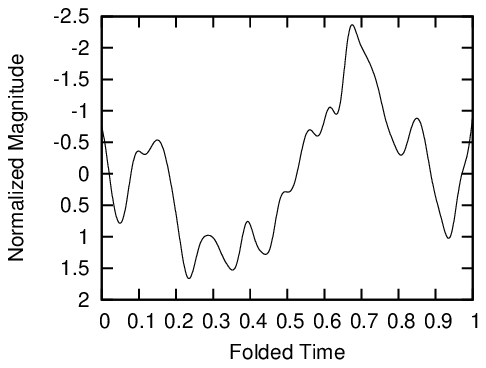} %
      } %
      \subfigure[13.203944 (CEPH)] {
        \label{fig:13.203944}
	\includegraphics{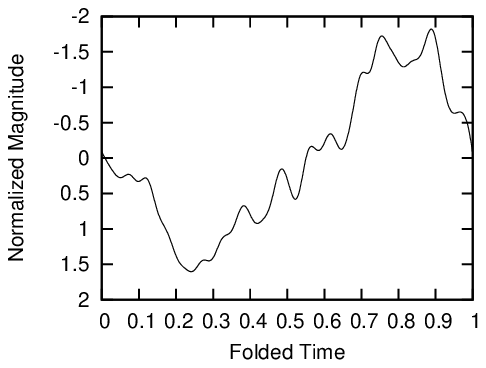} %
      } %

    }
    \mbox {
      \subfigure[11.331601 (CEPH)] {
	\label{fig:11.331601}
	\includegraphics{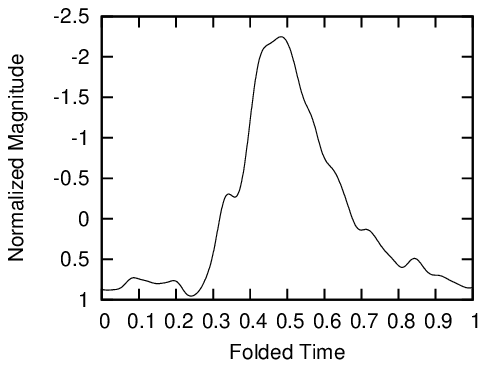} %
      } %

      \subfigure[21.40876 (CEPH)] {
	\label{fig:21.40876}
	\includegraphics{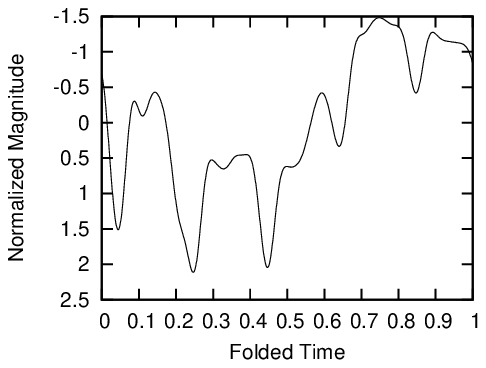} %
      } %
    }
    \caption{Selected anomalies (light-curve name and class appear below each figure). Note that the pre-processed
    versions of these light-curves are shown for purposes of clarity.}
    \label{fig:pavlos-anomalies}
\end{figure}

\begin{figure}[tbp]
    \centering
    \mbox {
      \subfigure[051657.87-690328.1 (EB)] {
	\label{fig:051657.87-690328.1}
	   \includegraphics{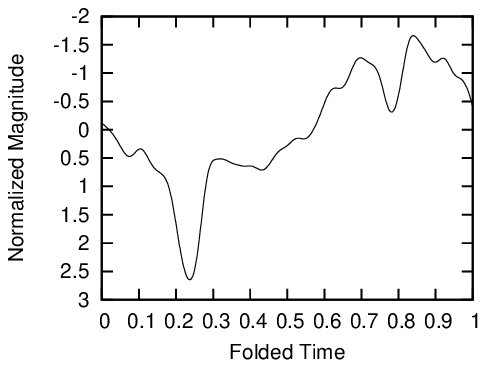} %
      } %

      \subfigure[053803.42-695656.4 (RRL)] {
	\label{fig:053803.42-695656.4}
	   \includegraphics{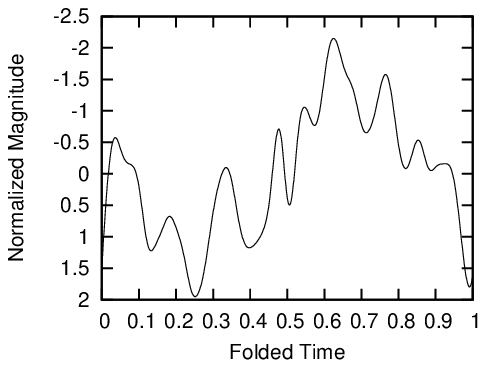} %
      } %
    }
    \mbox {
      \subfigure[052447.86-694319.0 (RRL)] {
	\label{fig:052447.86-694319.0}
	   \includegraphics{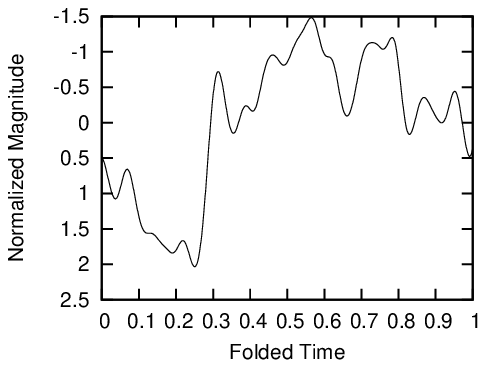} %
      } %
    }
    \caption{Selected anomalies (continued). Note that the pre-processed
    versions of these light-curves are shown for purposes of clarity.}
    \label{fig:pavlos-anomalies2}

\end{figure}

We conclude our experimental section with an analysis of the anomalies
output by PCAD by the astrophysicists on our project (Protopapas and
Alcock).  For the data sets CEPH, EB and RRL, they reviewed PCAD's ten
most anomalous light-curves and categorized the anomalies as follows:
a) noisy light-curves with high observational errors, b) light-curves
that were misclassified (e.g., a light-curve in the CEPH data set that
was not a Cepheid), and c) interesting examples worthy of further
study. Figures \ref{fig:pavlos-anomalies} and
\ref{fig:pavlos-anomalies2} show some of the more interesting
anomalies.  Please note that we show the pre-processed version of each
light-curve for clarity's sake, but our astrophysicists had the
original folded light-curves as well as other information at their
disposal.


We find that PCAD's top anomalies from the CEPH data set have few true
anomalies. The light-curve in Figure \ref{fig:17.70123} is an example
of a light-curve that lacks a strong period and has high observational
errors.  As it stands, its slow rise and fast decline is atypical of
Cepheids, but its measurements are too noisy to definitively conclude
it is an anomaly.  The light-curves in Figures \ref{fig:13.184117},
\ref{fig:20.54206} and \ref{fig:13.203944} exhibit the wrong asymmetry
and are most likely incorrectly classified as Cepheids. The
light-curve
\ref{fig:11.331601} has a long plateau which is atypical of Cepheid
stars, and is most likely not a Cepheid.  The light-curve in Figure
\ref{fig:21.40876} is interesting. The overall shape, period and color
(which are information accessible to the astrophysicists) are consistent
with a Cepheid.  However, its regularly spaced spikes cannot be
written off as noise, and may be indicative of an underlying dynamical
process.  A more careful study of this light-curve is needed to
understand its underlying physical process.

PCAD's top EB anomalies all have eccentric orbits, and are justifiably
flagged as anomalies. Only the light-curve in Figure
\ref{fig:051657.87-690328.1} is not a typical Eclipsing Binary. There
is either a third body present in the system producing a second
occultation or some form of atmospheric variation in one of the stars
in the system. There also might be a large reflection effect, which
occurs when the side of the dimmer star that is facing the Earth is
illuminated by the brighter companion star, thus increasing the
luminosity of the system \citep{pollacco93new}. This light curve also
warrants further investigation.


Among the RRL light-curves, PCAD identified two light curves that
are most likely misclassified. The light-curve in Figure
\ref{fig:053803.42-695656.4} has a quoted period and amplitude that
does not correspond to a typical RR Lyrae.  Furthermore, it may not
even be a periodic star. The light-curve in Figure
\ref{fig:052447.86-694319.0} does not have a strong signal and has a
long plateau which is atypical of RRL.  The rest of the anomalies are
very faint objects that have the characteristic shape of a RR Lyrae,
but a low signal-to-noise ratio that causes them to appear different
from the rest of the data set.

\section{Conclusion and Future Work}

This paper establishes PCAD as a method that discovers distance-based
local and global anomalies on unsynchronized periodic time series
data.  We use unsupervised learning to generate a set of
representative time series centroids from which the anomaly scores are
calculated.  Our method is able to scale to large data sets through
the use of sampling.  The online portion of our method is linear in
the size of the data.

In future work, we wish to develop density-based anomaly scores and
ranking methods that are robust to variance within a data set. We also
wish to validate our technique on larger amounts of light-curve data.
Specifically, our astrophysics collaborators will soon have access to
billions of light-curves through the Pan-STARRS survey.  This data
will be the ultimate test of PCAD's scalability.  We also wish to
incorporate the observational errors of the light-curves into the PCAD
algorithm, in order to fully automate the anomaly detection process.


\begin{acknowledgement}
The authors would like to acknowledge Dr. Anselm Blumer and Dr. Roni
Khardon of Tufts University for their helpful guidance and
comments. We also acknowledge Dr. Mark Friedl of Boston University for
the donation of the landcover data set, and Dr. David Walt and Matthew
Aernecke of Tufts University for the donation of the nose data
set. This material is based upon work supported by NASA under Grant
No. NNX07AV75G, and the NSF under Grant No. 0713259.
\end{acknowledgement}

\bibliographystyle{plainnat}
\bibliography{ml.resub.R3}

\end{document}